\documentclass[sigconf, nonacm, pdfa]{acmart}

\usepackage{xcolor}
\usepackage{xspace}
\usepackage{comment}
\usepackage{booktabs}
\usepackage{listings}
\usepackage{makecell}
\usepackage{etoolbox}
\usepackage{url}
\usepackage{hyperref}
\usepackage{amsmath,amsfonts}
\usepackage{algorithm}
\usepackage{algpseudocode}
\usepackage{graphicx}
\usepackage{textcomp}
\usepackage{tikz}
\usepackage{caption}
\usepackage{subcaption}
\usepackage{balance}
\usepackage[a-2b]{pdfx}

\usepackage{enumitem}
\setlist{nolistsep}

\DeclareRobustCommand\circled[1]{\tikz[baseline=(char.base)]{
    \node[shape=circle,draw,fill=white,inner sep=1pt] (char)
    {\textcolor{black}{#1}};}}

\newtoggle{showmarks}
\toggletrue{showmarks}

\iftoggle{showmarks}{
  \newcommand\myzhao[1]{\textcolor{violet}{[~MARK:~#1~]}}
  \newcommand\todo[1]{\textcolor{red}{[~TODO:~#1~]}}
  \newcommand\remark[1]{\textcolor{orange}{[~REMARK:~#1~]}}
  \newcommand\christos[1]{\textcolor{blue}{[~Christos:~#1~]}}
}{
  \newcommand\myzhao[1]{\unskip}
  \newcommand\todo[1]{\unskip}
  \newcommand\remark[1]{\unskip}
  \newcommand\christos[1]{\unskip}
}

\newcommand\rev[1]{\textcolor{black}{#1}}
\newcommand\shep[1]{\textcolor{black}{#1}}

\newcommand{\systemname}{\textit{cedar}\xspace}

\definecolor{codegreen}{rgb}{0,0.6,0}
\definecolor{codegray}{rgb}{0.5,0.5,0.5}
\definecolor{codepurple}{rgb}{0.58,0,0.82}
\definecolor{backcolour}{rgb}{0.95,0.95,0.92}

\lstdefinestyle{python}{
    backgroundcolor=\color{backcolour},   
    commentstyle=\color{codegreen},
    keywordstyle=\color{magenta},
    numberstyle=\tiny\color{codegray},
    stringstyle=\color{codepurple},
    basicstyle=\ttfamily\footnotesize,
    breakatwhitespace=false,         
    breaklines=true,                 
    captionpos=b,                    
    keepspaces=true,                 
    numbers=left,                    
    numbersep=5pt,                  
    showspaces=false,                
    showstringspaces=false,
    showtabs=false,                  
    tabsize=2
}

\newcommand\vldbdoi{10.14778/3705829.3705861}
\newcommand\vldbpages{488 - 502}
\newcommand\vldbvolume{18}
\newcommand\vldbissue{2}
\newcommand\vldbyear{2024}
\newcommand\vldbauthors{\authors}
\newcommand\vldbtitle{\shorttitle} 
\newcommand\vldbavailabilityurl{https://github.com/stanford-mast/cedar}
\newcommand\vldbpagestyle{empty} 

\begin{document}
\title{\systemname: Optimized and Unified Machine Learning Input Data Pipelines}

\author{Mark Zhao}
\affiliation{%
  \institution{Stanford University}
}
\email{myzhao@cs.stanford.edu}

\author{Emanuel Adamiak}
\affiliation{%
  \institution{Stanford University}
}
\email{adamiak@stanford.edu}

\author{Christos Kozyrakis}
\affiliation{%
  \institution{Stanford University}
}
\email{christos@cs.stanford.edu}

\begin{abstract}
The input data pipeline is an essential component of each machine learning (ML) training job.
It is responsible for reading massive amounts of training data, processing batches of samples using complex transformations, and loading them onto training nodes at low latency and high throughput.
Performant input data systems are becoming increasingly critical due to skyrocketing data volumes and training throughput demands.
Unfortunately, current input data systems cannot fully leverage key performance optimizations, resulting in hugely inefficient infrastructures that require significant resources -- or worse -- underutilize expensive accelerators.

To address these demands, we present \systemname, an optimized and unified programming framework for ML input data pipelines.
\systemname allows users to define a training job's data pipeline using composable operators that support arbitrary ML frameworks and libraries.
\systemname's extensible optimizer systematically combines and applies performance optimizations to the pipeline.
\systemname then orchestrates pipeline processing across configurable local and distributed compute resources to efficiently meet the training job's data throughput demands.
\rev{Across eight pipelines, \systemname improves performance by up to $1.87\times$ to $10.65\times$ compared to state-of-the-art input data systems.}
\end{abstract}

\maketitle

\pagestyle{\vldbpagestyle}
\begingroup\small\noindent\raggedright\textbf{PVLDB Reference Format:}\\
\vldbauthors. \vldbtitle. PVLDB, \vldbvolume(\vldbissue): \vldbpages, \vldbyear.\\
\href{https://doi.org/\vldbdoi}{doi:\vldbdoi}
\endgroup
\begingroup
\renewcommand\thefootnote{}\footnote{\noindent
This work is licensed under the Creative Commons BY-NC-ND 4.0 International License. Visit \url{https://creativecommons.org/licenses/by-nc-nd/4.0/} to view a copy of this license. For any use beyond those covered by this license, obtain permission by emailing \href{mailto:info@vldb.org}{info@vldb.org}. Copyright is held by the owner/author(s). Publication rights licensed to the VLDB Endowment. \\
\raggedright Proceedings of the VLDB Endowment, Vol. \vldbvolume, No. \vldbissue\ %
ISSN 2150-8097. \\
\href{https://doi.org/\vldbdoi}{doi:\vldbdoi} \\
}\addtocounter{footnote}{-1}\endgroup

\ifdefempty{\vldbavailabilityurl}{}{
\vspace{.3cm}
\begingroup\small\noindent\raggedright\textbf{PVLDB Artifact Availability:}\\
The source code, data, and/or other artifacts have been made available at \url{\vldbavailabilityurl}.
\endgroup
}

\section{Introduction}\label{sec:introduction}
Every deep machine learning (ML) training job relies on an \textit{input data pipeline} to transform raw datasets into prepared training samples (i.e., mini-batches of tensors), ready to be consumed by the ML framework (e.g., PyTorch~\cite{nips19:paszke_pytorch} or TensorFlow~\cite{whitepaper:tensorflow}).
These pipelines are executed by \textit{input data systems} such as PyTorch's DataLoader~\cite{website:torch-utilsdata} or TensorFlow's tf.data~\cite{vldb14:murray_tfdata}.
Input data systems are a distinct and essential component of the end-to-end ML data pipeline, complementing the traditional parallel processing frameworks, such as Spark~\cite{nsdi12:zaharia_spark} and Beam~\cite{vldb8:akidau_dataflow}, commonly used for {\it offline} ML data ingestion.
In contrast, input data pipelines require \textit{online} preprocessing using diverse domain- and ML framework-specific tensor operations that vary heavily across training jobs.
Input data systems are designed to execute these pipelines while meeting the stringent performance requirements~\cite{isca22:zhao_dsi, vldb14:mohan_coordl, socc23:audibert_tfdataservice} of each training job.

For example, a computer vision (CV) input data pipeline may use various Python libraries and UDFs (e.g., OpenCV~\cite{opencv} or torchvision~\cite{website:torchvision}) to decode JPEG images from a dataset, apply random augmentations such as crops and distortions to each image, and convert batches of images into a tensor. %
Each training job uses an input data system to continuously execute the input data pipeline, potentially over multiple epochs, throughout the training job's lifetime.
The input data system must carefully match its throughput to accelerator demands.
This avoids data stalls~\cite{vldb14:mohan_coordl}, which degrade training throughput, without over-provisioning input data resources.

Recently, the input data throughput required by training jobs has grown at an immense rate~\cite{isca22:zhao_dsi, sigmod23:zhao_goldminer, vldb14:mohan_coordl, vldb14:murray_tfdata}, driven by specialized hardware~\cite{website:nvidia-mlperf, website:aws_trainium, isca23:jouppi_tpuv4}, optimized software techniques~\cite{nips15:han_learning, iclr16:han_deepcompression}, and massive training clusters~\cite{website:tpuv4-cloud, isca22:mudigere_zionex, website:rsc, nsdi24:jiang_megascale, website:meta-genai}.
To avoid costly data stalls, companies such as Google~\cite{socc23:audibert_tfdataservice}, Alibaba~\cite{sigmod23:zhao_goldminer}, and Meta~\cite{isca22:zhao_dsi} have deployed distributed input data services.
While simply scaling-out compute addresses performance bottlenecks, it comes with an immense resource cost.
Meta's DPP can require dozens of CPU servers to support a \textit{single} GPU server~\cite{isca22:zhao_dsi}, while a single model at Google can require more than five thousand preprocessing workers~\cite{socc23:audibert_tfdataservice}!
To continue scaling ML infrastructure, \textit{it is critical to optimize both the performance and efficiency of input data systems}.

While recent research has explored various performance optimizations such as caching~\cite{atc22:graur_cachew, eurosys23:zhao_silod, vldb14:mohan_coordl, fast20:kumar_quiver, sigmod22:isenko_presto, hpca23:chen_icache, fast23:khan_shade, atc23:zhao_tectonicshift}, offloading to high-performance backends~\cite{vldb14:murray_tfdata, vldb16:um_fastflow, sigmod23:zhao_goldminer, atc22:graur_cachew, mlsys22:kuchnik_plumber, isca22:zhao_dsi}, prefetching~\cite{vldb14:murray_tfdata}, and fusion~\cite{vldb14:murray_tfdata}, current input data systems are insufficient for several reasons.
First, current systems apply optimizations in an \textit{isolated} manner using bespoke and \textit{inextensible} solutions.
Combining multiple optimizations is crucial; however these systems cannot navigate the complex search space that is needed to enable this combination. %
Thus, users are currently forced to pick and choose optimizations, sacrificing performance and efficiency.
Secondly, current systems are \textit{not context-aware}, precluding key optimizations that require understanding application semantics. %
Finally, current systems are \textit{specialized} -- they cannot support the wide breadth of ML frameworks, domain-specific libraries, and execution engines in use across the ML training landscape.
For example, many input data systems, including tf.data~\cite{vldb14:murray_tfdata}, tf.data service~\cite{socc23:audibert_tfdataservice}, Cachew~\cite{atc22:graur_cachew}, Plumber~\cite{mlsys22:kuchnik_plumber}, FastFlow~\cite{vldb16:um_fastflow}, PRESTO~\cite{sigmod22:isenko_presto}, and GoldMiner~\cite{sigmod23:zhao_goldminer} rely on TensorFlow's dataflow graph and execution backend, limiting their compatibility with non-TensorFlow frameworks such as PyTorch.
This is especially worrisome given TensorFlow's decline among ML practitioners -- \textit{supporting only 2\% of recent ML research} as of September 2024~\cite{website:paperswithcode-trends}.

\begin{figure}[t]
  \centering
  \includegraphics[width=\linewidth]{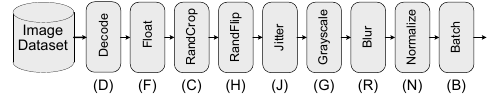}
  \caption{A computer vision input data pipeline applies a sequence of augmentations to each image.}
  \label{fig:simclr}
\end{figure}

\noindent\textbf{A Motivating Example.}
To illustrate these limitations, consider the challenges faced by an ML practitioner training a PyTorch CV model.
The practitioner uses the input data pipeline shown in Figure~\ref{fig:simclr} to apply augmentations~\cite{arxiv:chen_simclr, arxiv:chen_simclrv2} to each image.
Suppose during testing, the practitioner discovers a data stall~\cite{vldb14:mohan_coordl}.
The practitioner recalls some optimizations, such as caching, using distributed workers, reordering operators, and fusion, and begins manually testing different execution plans.
They quickly realize both the complexities in applying a single optimization (discovering that caching can \textit{harm} throughput), and the difficulties in combining them (resolving an ideal operator fusion that conflicts with an ideal operator ordering)\footnote{We experimentally illustrate these scenarios in Section~\ref{sec:optimizations}.}.
Upon realizing the billions of execution plans they need to manually implement and test, they decide to port their entire model and pipeline to TensorFlow to leverage its distributed input data frameworks, such as tf.data service~\cite{socc23:audibert_tfdataservice} or FastFlow~\cite{vldb16:um_fastflow}. 
Upon observing that the disaggregated backend needs further optimizations to improve resource efficiency, the practitioner is forced to reluctantly continue manually exploring optimizations. %

To address this gap, we believe there is a strong need for an input data framework that can systematically apply a suite of system optimizations and navigate the complex search space that these optimizations introduce. %
Such a framework must address the unique challenges of ML input data pipelines, which render the straightforward application of traditional query optimizers ineffective.
These challenges include the need to support diverse ML frameworks, and the ability to optimize pipelines -- programmed via opaque Python UDFs -- that heavily rely on domain-specific preprocessing libraries and execution engines.
The application of these optimizations must also be \textit{context-aware}, respecting requirements (e.g., randomness) and leveraging novel optimization opportunities (e.g., semantic-preserving reorderings\footnote{e.g., applying blur and crop to an image in either order yields the same semantics.}) presented by input data pipelines.

\noindent\textbf{Our Solution.}
We present \systemname, an optimized and unified Python-native programming framework for ML input data pipelines.
\systemname transparently enables \textit{systematic}, \textit{context-aware}, and \textit{general} optimizations to meet the high performance and efficiency needs of modern ML training systems.
\systemname allows ML practitioners to easily build input data pipelines by linking together modular native and higher-order operators functionally.
These pipelines can support a wide breadth of ML frameworks (e.g., PyTorch and TensorFlow) and domain-specific preprocessing libraries.
Meanwhile, \systemname automatically optimizes and manages the execution of the pipeline to meet the training job's throughput requirements, eliminating data stalls with high resource efficiency.
To do so, \systemname introduces an extensible Optimizer, which systematically applies a combination of state-of-the-art and novel context-aware input data optimizations to \rev{improve} throughput on a per-resource basis.
Importantly, to maintain the semantic correctness of black-box UDFs under these optimizations, the Optimizer leverages a set of \rev{simple} yet expressive hints specified by the practitioner that provide essential domain knowledge to \systemname.
Then during runtime, \systemname dynamically orchestrates processing across an extensible and scalable set of execution engines, such as a distributed cluster or local process pools on the training nodes' CPUs, according to the optimized plan.
\systemname continuously monitors and right-sizes resources, efficiently meeting the training job's throughput demands.

We evaluated \systemname on a diverse set of \rev{eight} input data pipelines across ML domains, using both local and distributed execution engines.
By understanding the complex systems dynamics that arise in each pipeline, \systemname successfully leverages a mix of optimizations and engines to \rev{improve} each pipeline's throughput.
\rev{
\systemname outperforms state-of-the-art ML input data systems, including tf.data~\cite{vldb14:murray_tfdata}, tf.data service~\cite{socc23:audibert_tfdataservice}, FastFlow~\cite{vldb16:um_fastflow}, Plumber~\cite{mlsys22:kuchnik_plumber}, Ray Data~\cite{website:ray-data}, and PyTorch's DataLoader~\cite{website:torch-utilsdata} by up to $1.87\times$ to $10.65\times$ while utilizing the same set of resources. 
}
\systemname then effectively \rev{scales} resource allocations to meet diverse training throughput demands, translating this performance benefit to high input data system efficiency.

In summary, we make the following contributions.
\begin{itemize}[leftmargin=*]
    \item We introduce an extensible optimization framework for ML input data pipelines that automatically explores the massive combinatorial search space introduced by concurrent optimizations. 
    \item We introduce easy-to-use interfaces that enable novel input data optimizations (i.e., semantic-preserving reorderings) that rely on the domain knowledge of black-box UDFs.
    \item We present \systemname, a unified input data framework that transparently optimizes and orchestrates pipeline processing, supporting a wide range of ML frameworks and execution engines. %
\end{itemize}

\systemname provides an important, yet missing, foundation for input data systems research, analogous to the influence extensible optimizers (e.g., Spark SQL's Catalyst~\cite{sigmod15:armburst_sparksql}) and execution interfaces (e.g., Beam's Runners~\cite{website:beam}) have had in traditional data processing.

\section{ML Data Ingestion Background}\label{sec:background}
\begin{figure}[t]
  \centering
  \includegraphics[width=\linewidth]{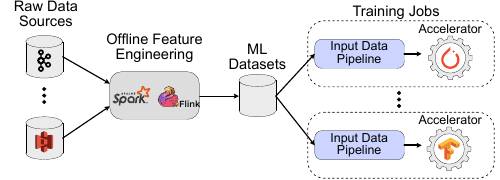}
  \caption{ML training data pipelines consist of an offline feature engineering and an online input data processing stage.}
  \label{fig:background_pipeline}
\end{figure}

ML training jobs rely on data ingestion pipelines to transform raw operational data into structured samples (i.e., tensors) interpretable by ML frameworks such as PyTorch~\cite{nips19:paszke_pytorch} and TensorFlow~\cite{whitepaper:tensorflow}.
As shown in Figure~\ref{fig:background_pipeline}, these pipelines traditionally consist of two phases: offline \textit{feature engineering} and online \textit{input data processing}.

\noindent\textbf{Offline Feature Engineering.}
Feature engineering pipelines validate, aggregate, join, and transform raw operational data into structured \textit{datasets}, off the critical path of training.
Since feature engineering predominantly requires traditional extract-transform-load (ETL) tasks, ML practitioners commonly use general-purpose distributed processing frameworks such as Spark~\cite{nsdi12:zaharia_spark}, Flink~\cite{tcde15:katsifodimos_flink}, and Beam~\cite{vldb8:akidau_dataflow}. %
These ETL tasks are independent from training jobs.
They run prior to training and materialize datasets that are stored across a variety of systems, from training nodes' local file systems to distributed data lakes~\cite{atc23:zhao_tectonicshift, fast21:pan_tectonic, vldbv13:armburst_deltalake, sigmod16:dageville_snowflake}.

\noindent\textbf{Online Input Data Pipelines.}
Each training job concurrently runs an input data pipeline to perform the ``last-mile'' of processing -- extracting a job-specific subset of training samples from stored datasets, transforming them into ready-to-use mini-batches of tensors, and loading tensors into the host memory of the training node -- all on-the-fly.
This ``online'' processing is needed because operations often vary heavily across jobs or even across epochs within the same job, making materialization highly inefficient.
For example, NLP practitioners often experiment with tokenization algorithms~\cite{website:huggingface-tokenizers} across model architectures.
Recommendation models require different hashing configurations depending on their embedding table dimensions~\cite{asplos22:sethi_recshard}.
Many domains, including CV~\cite{neurips20:cubuk_randaugment}, NLP~\cite{arxiv:feng_nlp-augmentation}, and speech~\cite{interspeech19:park_specaugment}, apply random augmentations that re-process each sample every epoch to improve model generalization.

As a result, input data pipelines have requirements distinct from traditional ETL tasks, demanding a dedicated class of systems.
Input data systems must right-size resources and generate each mini-batch to meet strict low latency ($O(100\mu s - 1ms)$ per step)~\cite{socc23:audibert_tfdataservice} and high throughput ($O(10 GB/s)$ per node)~\cite{isca22:zhao_dsi} requirements throughout the lifetime of the job, which can take days to weeks.
Violations of these requirements can bottleneck expensive accelerators~\cite{vldb14:mohan_coordl}.
Furthermore, ML input data pipelines predominantly operate on unstructured tensors within a mini-batch granularity and rely heavily on myriad domain-specific Python libraries and UDFs.
For example, practitioners may call a HuggingFace~\cite{website:huggingface-tokenizers} tokenizer or a torchvision~\cite{website:torchvision} crop for each text or image training sample, respectively.

Distributed input data systems, such as Meta's DPP~\cite{isca22:zhao_dsi} and Google's tf.data service~\cite{socc23:audibert_tfdataservice}, are increasingly deployed to address throughput demands.
\rev{However, they present an immense resource cost that constrains the scalability of ML training systems.
For example, tf.data service deployments at Google commonly use between 2 and 32 (and up to 5K) distributed workers for each training job~\cite{socc23:audibert_tfdataservice}. 
Recommendation training jobs at Meta can require dozens of DPP workers for \textit{each} GPU training node --  demanding comparable power capacity to the training node itself~\cite{isca22:zhao_dsi}.}

\section{Optimizing Input Data Pipelines}\label{sec:optimizations}

\subsection{Requirements}\label{sec:optimizations_requirements}
Optimizing the performance and efficiency of input data systems is essential.
Recent systems have begun to explore techniques such as parallelism and disaggregation~\cite{vldb14:murray_tfdata, vldb16:um_fastflow, sigmod23:zhao_goldminer, atc22:graur_cachew, mlsys22:kuchnik_plumber, isca22:zhao_dsi}, caching~\cite{atc22:graur_cachew, eurosys23:zhao_silod, vldb14:mohan_coordl, fast20:kumar_quiver, sigmod22:isenko_presto, hpca23:chen_icache, fast23:khan_shade, atc23:zhao_tectonicshift, atc21:lee_refurbish}, operator fusion~\cite{website:ray-data, vldb14:murray_tfdata}, and inter-job coordination~\cite{hotcloud19:kakaraparthy_oneaccess, vldb14:mohan_coordl, atc22:graur_cachew, eurosys23:zhao_silod, fast20:kumar_quiver}.
Unfortunately, current input data systems cannot enable the \textit{systematic}, \textit{context-aware}, and \textit{general} optimizations critical to achieving this goal.

\noindent\textbf{Need for Systematic Optimizations.} As we deeply explore in Section~\ref{sec:optimizations_space}, myriad input data optimizations (e.g., offloading, fusion, caching, reordering, and prefetching) can be effective at significantly improving performance.
Combining these optimizations is essential, improving throughput by $4.44\times$ versus a single technique in isolation.
To do so, systems need a deep understanding of the interactions between optimizations and a systematic approach to navigating the complex tradeoffs and search space introduced.

Current systems cannot systematically combine and apply optimizations.
They lack the notion of an extensible optimizer and instead craft bespoke solutions focused only on a limited and isolated set of techniques.
For example, Plumber~\cite{mlsys22:kuchnik_plumber} uses a complex linear program to tune operator parallelism, but its formulation is constrained to a single processing node.
Meanwhile, FastFlow~\cite{vldb16:um_fastflow} can leverage distributed workers for processing, but explicitly chooses between only three execution plans, precluding additional optimizations such as caching intermediate outputs or operator fusion.
PRESTO~\cite{sigmod22:isenko_presto} introduces a profiler that can inform certain optimizations, such as the optimal location to cache, but requires users to manually reason about and perform these changes.

\noindent\textbf{Need for Context-Aware Optimizations.}
Unlike traditional data processing applications with standardized APIs, like SQL, input data pipelines heavily rely on opaque Python UDFs with unique operational semantics.
This can both complicate and serve as an opportunity for optimizations.
For example, caching after certain stochastic operators may harm model convergence~\cite{atc21:lee_refurbish}.
Meanwhile, certain reorderings may yield performance benefits while preserving intended semantics; for example performing a crop before a blur eliminates wasted work. %
Input data systems require domain knowledge for these context-specific considerations.

Various input data systems such as tf.data~\cite{vldb14:murray_tfdata} and TorchData~\cite{website:torchdata} allow users to build structured pipelines from these UDFs via higher-order operators (e.g., \textit{map} and \textit{filter}).
However, these systems still require users to manually reason about and perform optimizations, such as explicitly inserting a 
\textit{cache} operator in the pipeline. 
While systems like Cachew~\cite{atc22:graur_cachew} can automatically identify optimal caching points, they require users to explicitly indicate where caching is allowed.
By requiring users to directly modify the pipeline, such interfaces limit the integration of further optimizations.
For example, with reordering, users would need to modify the combinatorial set of orderings with permissible cache locations.
Ideally, input data systems should enable users to provide simple yet expressive hints that facilitate context-aware optimizations and ensure correctness, but without precluding further optimizations.

\noindent\textbf{Need for General Optimizations.} 
Finally, ML practitioners use a variety of preprocessing libraries and ML frameworks tailored to their needs.
For example, an NLP practitioner may use HuggingFace's Tokenizer library~\cite{website:huggingface-tokenizers}, while an ASR practitioner may require MP3 decoding methods from librosa~\cite{scipy15:mcfee_librosa}.
Furthermore, these libraries often rely on different execution backends, such as Apache Arrow~\cite{website:arrow} or TensorFlow kernels.
These input data pipelines then supply data to various ML training platforms, such as PyTorch~\cite{nips19:paszke_pytorch}, Jax~\cite{website:jax}, TensorFlow~\cite{whitepaper:tensorflow}, MindSpore~\cite{website:mindspore}, and others~\cite{website:paperswithcode-trends}.
Input data systems should be able to apply optimizations across diverse domain-specific libraries, execution engines, and ML frameworks.

Unfortunately, many systems restrict their scope to optimizing a subset of use cases. %
For example, many (if not most) input data systems, including PRESTO~\cite{sigmod22:isenko_presto}, Cachew~\cite{atc22:graur_cachew}, tf.data service~\cite{socc23:audibert_tfdataservice}, tf.data~\cite{vldb14:murray_tfdata}, GoldMiner~\cite{sigmod23:zhao_goldminer}, FastFlow~\cite{vldb16:um_fastflow}, and Plumber~\cite{mlsys22:kuchnik_plumber} are built on top of TensorFlow's static graph abstraction.
While doing so allows these systems to gain the benefits of TensorFlow's graph optimizations~\cite{website:tf-graph}, this severely limits their applicability to the limited set of practitioners who rely on TensorFlow~\cite{website:paperswithcode-trends}.
For example, these systems cannot optimize pipelines that rely on third-party training frameworks or libraries.
In a similar vein, other recent works use specific execution engines for input data processing (e.g., Ray~\cite{osdi18:moritz_ray} with Ray Data~\cite{website:ray-data}, or GPUs with FusionFlow~\cite{vldb17:kim_fusionflow} and DALI~\cite{website:dali}).
While it is critical to leverage the benefits of these specialized engines, it is also important to not be \textit{limited} to them.

Finally, we note that recent works also address additional considerations, such as multi-tenant environments.
For example, Cachew \cite{atc22:graur_cachew}, CoorDL~\cite{vldb14:mohan_coordl}, Quiver~\cite{fast20:kumar_quiver}, OneAccess~\cite{hotcloud19:kakaraparthy_oneaccess}, and Tectonic-Shift \cite{atc23:zhao_tectonicshift} are designed to exploit data reuse in multi-tenant scenarios where datasets and transformations are shared across concurrent training jobs.
In this paper, we focus on optimizing the performance and efficiency of input data systems for a single training job.
However, because these works leverage input data systems as a foundation (e.g., Cachew builds on tf.data), our optimizations will be essential to the performance of these orthogonal applications.

\subsection{The Complex Optimization Space}\label{sec:optimizations_space}
\rev{We begin by analyzing and distilling insights behind an extensive set of optimization techniques that are currently applied in an isolated manner: offloading, fusion, caching, and prefetching.
We also introduce a new optimization in the context of input data pipelines: semantic-preserving reorders.
}
We used the representative CV pipeline\footnote{We refer to each operator by its letter abbreviation in this section.} for SimCLR~\cite{arxiv:chen_simclr, arxiv:chen_simclrv2} shown in Figure~\ref{fig:simclr}, and we performed experiments on an 8-core (n2-standard-8) and 32-core (n2-standard-32) VM on Google Cloud. %
The pipeline reads a subset of the ImageNet dataset~\cite{cvpr09:deng_imagenet} stored on the file system of the 8-core VM.
Unless otherwise specified, each optimization extends a \texttt{baseline} that executes all operators in the main data loading process on the 8-core VM.
\rev{We focus on how these optimizations improve raw input data throughput given a fixed amount of resources, which translates to improved training throughput and efficiency.} %

\begin{figure}[t]
  \centering
  \includegraphics[width=\linewidth]{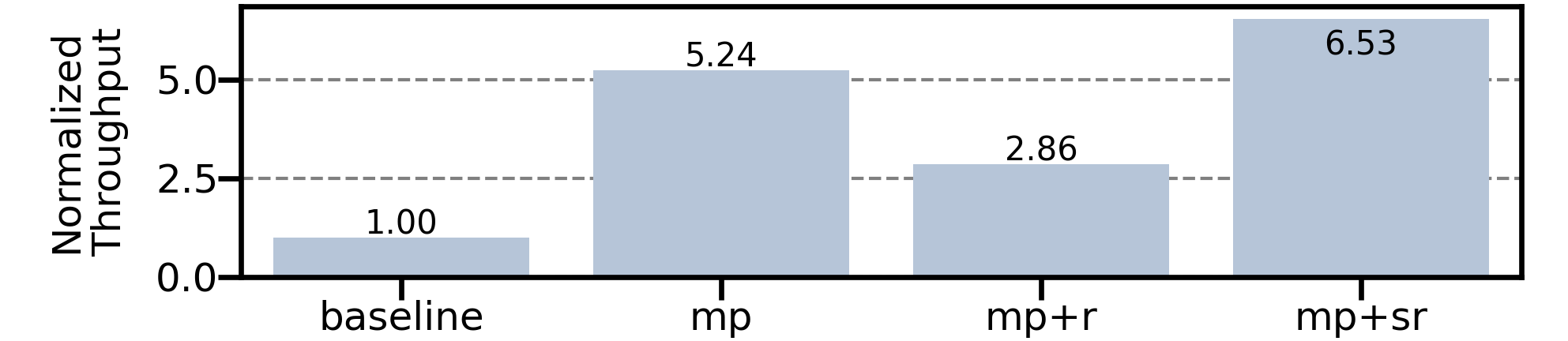} 
  \caption{\rev{Pipeline throughput, normalized to the baseline, by offloading operators across local and remote backends.}}
  \label{fig:offloading}
\end{figure}

\noindent\textbf{Offloading.} %
\rev{Many input data systems offload the entire pipeline to parallel execution engines (e.g., thread/process pools~\cite{website:torch-utilsdata, vldb14:murray_tfdata} or distributed workers~\cite{isca22:zhao_dsi, sigmod23:zhao_goldminer, socc23:audibert_tfdataservice}).
The choice of execution engine presents complex tradeoffs on a pipeline and per-operator basis.}
Figure~\ref{fig:offloading} shows the \rev{throughput} of the \texttt{CV} pipeline across various engines, normalized to the \texttt{baseline}. %
\texttt{mp} offloads execution of the entire pipeline to a local multiprocess pool.
\texttt{mp+r} offloads the entire pipeline to multiple processes on the remote 32-core VM, with local processes facilitating RPCs.
Surprisingly, \textit{more} parallelism \textit{harms} \rev{performance} because each local process performs more work communicating than it would performing the operators themselves.
That being said, remote backends \textit{can} offer benefits if used correctly.
\texttt{mp+sr} only \textit{selectively} offloads the blur operator, which exhibits a high arithmetic intensity, improving \rev{throughput} by $25\%$ over \texttt{mp}. 

\rev{
\textit{Insight.}
Each operator exhibits diverse performance benefits, \textit{or losses}, across engines.
Carefully select \textit{each} operator's engine, as opposed to relying on a single engine for the entire pipeline.
}

\begin{figure}[t]
  \centering
  \includegraphics[width=\linewidth]{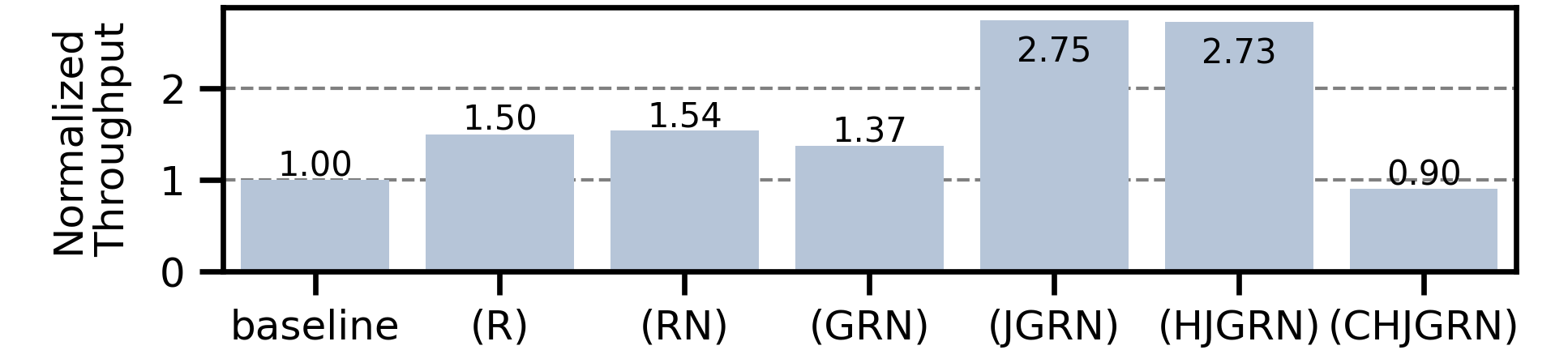} 
  \caption{\rev{Pipeline throughput, normalized to the baseline, by executing fused operators (see Figure~\ref{fig:simclr}) remotely.}}
  \label{fig:fusion}
\end{figure}

\noindent\textbf{Fusion.} %
\rev{Fusing operators can improve performance by eliminating intermediate data transfers.}
Figure~\ref{fig:fusion} shows the \rev{throughput} of the \texttt{CV} pipeline when fusing and offloading certain operators to the remote 32-core VM.
Fusing and offloading neighbors to blur (R) can either improve or \textit{degrade} performance.
Fusing grayscale (GRN) harms performance because doing so sends RGB images remotely, incurring I/O costs.
Meanwhile, continuing to fuse the compute-intensive jitter (JGRN) is optimal as it best leverages parallelism.
Further fusing crop (CHJGRN) eliminates these benefits, as full-sized images now need to be sent to the remote VM.

\rev{\textit{Insight.}
Simple heuristics (e.g., fusing all adjacent \texttt{maps}~\cite{vldb14:murray_tfdata}) has severe pitfalls.
Instead, fusions must be systematically applied based on each operator's dynamics (e.g., I/O and compute demands).
}

\begin{figure}[t]
  \centering
  \includegraphics[width=\linewidth]{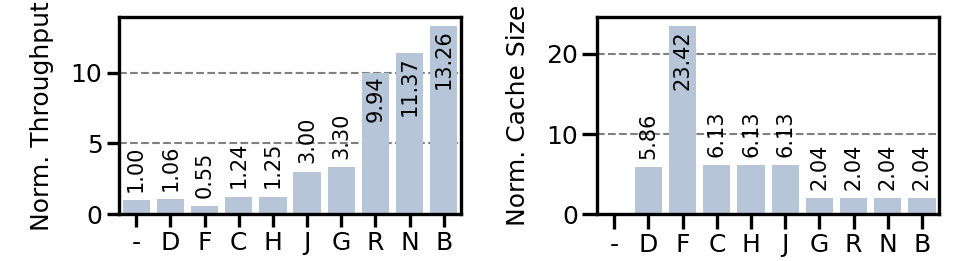} 
  \caption{\rev{Pipeline throughput and cache size requirement after materializing the output of each operator (see Figure~\ref{fig:simclr}), normalized to no caching and the raw dataset size (`-').}}
  \label{fig:caching}
\end{figure}

\noindent\textbf{Caching.} %
\rev{Caching can improve performance by trading off compute for storage costs.}
Figure~\ref{fig:caching} shows the execution time and intermediate cache size observed when caching the output of a specific operator to the disk of the 8-core VM. %
Caching can harm throughput, such as caching the output of int8 to fp32 conversion (F).
Doing so increases the size of each sample, incurring more I/O overheads relative to compute saved.
Furthermore, caching after any \textit{random} operator (e.g., crop (C)) would violate the stochastic semantics of the pipeline, harming the model's accuracy.
Considering these factors, caching the CV pipeline is largely ineffective without additional optimizations!
The best semantic-preserving cache location (after decode (D)) minimally reduces execution time at large storage cost.

\rev{
\textit{Insight.}
Caching must consider both random operators, and the relative savings in compute compared to storage and I/O costs.
}

\begin{figure}[t]
  \centering
  \includegraphics[width=\linewidth]{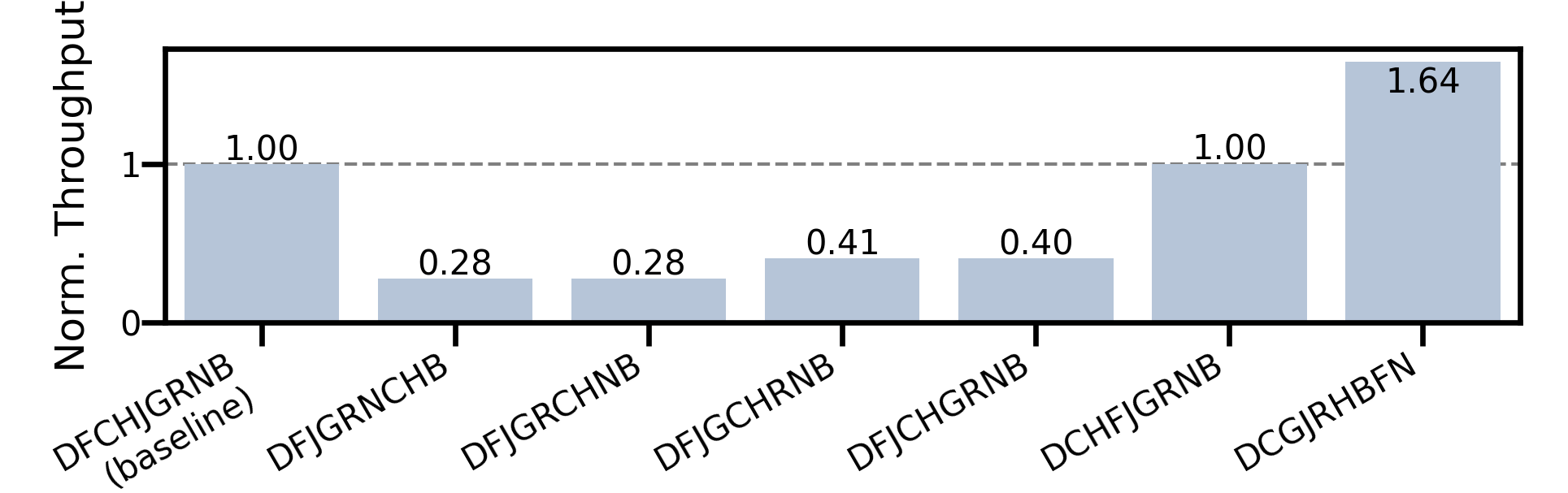} 
  \caption{\rev{Pipeline throughput (without offloading) by reordering various operators in the \texttt{CV} pipeline (see Figure~\ref{fig:simclr}).}}
  \label{fig:reordering}
\end{figure}

\noindent\textbf{Reordering.} %
While operator reordering is well-known in database systems~\cite{jvldb22:kossman_datadependencies}, its benefits have yet to be explored for input data pipelines.
Figure~\ref{fig:reordering} shows how there is a $5.90\times$ variation in execution time across seven operator orderings.
The ideal reordering pushed size-reducing transformations (e.g., crop (C) and grayscale (G)) towards the beginning and size-increasing transformations (e.g., int8 to fp32 conversion (F)) towards the output.
Note that certain reorderings can cause minor variations in the output (e.g., reordering a blur and crop) while preserving its overall semantics (a cropped, blurred image).
ML training jobs are robust to these variations (e.g., augmentations are often applied in random order~\cite{neurips20:cubuk_randaugment}).

\rev{\textit{Insight.}
Reordering operators based on how they change sample size can improve performance by reducing the compute required for each sample.
However, safely reordering operators requires domain knowledge from users to specify permissible reorderings.
}

\noindent\textbf{Prefetching.} %
Prefetching the output of the pipeline allows input data processing to be overlapped with the training step.
For example, we observed a $35\%$ improvement in end-to-end training throughput for the \texttt{CV} pipeline by prefetching its output, assuming a $100ms$ training step. %
Furthermore, prefetching the output of an offloaded operator can overlap its computation with downstream operators.
We observed a $30\%$ improvement in overall input data throughput by prefetching the output of an offloaded blur operator. %

\rev{\textit{Insight.} Input data systems should prefetch both the pipeline output and offloaded operators to overlap and pipeline computation.}

\subsection{Our Approach}
\rev{Despite their impact, we do not believe that these optimizations are exhaustive; we instead advocate for an extensible platform that can incorporate future optimizations, akin to optimizers in traditional data processing systems.
Furthermore, it is critical to \textit{combine} multiple optimizations together to maximize the performance and efficiency of input data systems.
As we later show in Figure~\ref{fig:ablation_eval}, concurrently applying the above optimizations achieves a $4.44\times$ higher throughput than using a single optimization (local offloading), and a $21.81\times$ higher throughput than the \texttt{baseline}.
However, combining even the above optimizations requires the exploration of a vast search space: $\sim85$ billion plans for the \texttt{CV} pipeline.
}

\rev{
To address these challenges and the key requirements in Section~\ref{sec:optimizations_requirements}, we argue that input data systems need a higher level of components and abstractions.
First, an extensible \textit{optimizer}, leveraging structured cost- and rule-based optimization passes, is needed to systematically explore the complex set of plans generated when combining multiple optimizations.
Secondly, an easy-to-use and expressive programming model is essential to support diverse pipelines and to capture the necessary domain knowledge to enable key context-aware optimizations.
Finally, well-defined interfaces, providing an easy-to-manage intermediate representation, is vital to enable future optimizations and execution engines.
}

\section{\systemname Framework}\label{sec:design}
To this end, \systemname introduces higher-level abstractions and components that allow ML practitioners to easily build, optimize, and execute input data pipelines.
\systemname provides native and higher-order operators which practitioners use to define input data pipelines that are \textit{general} -- supporting arbitrary ML frameworks and libraries -- and \textit{logical} -- abstracting away underlying processing details (e.g., which engine or how much parallelism to use).
Practitioners can optionally provide a lightweight set of hints to provide domain knowledge for \textit{context-aware} optimizations.
\shep{\systemname's \texttt{Optimizer} then statically applies the optimizations presented in Section~\ref{sec:optimizations_space} to yield an optimized execution plan that improves performance on a per-resource basis.
During runtime, \systemname's \texttt{Scaler} auto-scales resources to execute this plan, across configurable engines, to efficiently meet training throughput requirements.
}

\begin{figure}[t]
  \centering
  \includegraphics[width=\linewidth]{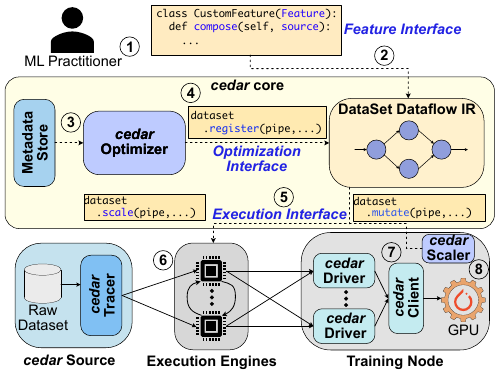}
  \caption{\rev{\systemname block diagram, showing how users can leverage \systemname to define, optimize, and execute pipelines.}}
  \label{fig:block_diagram}
\end{figure}

Figure~\ref{fig:block_diagram} shows an end-to-end example.
\rev{
\circled{1} A practitioner composes a \textit{logical} input data pipeline (a \texttt{Feature}) by functionally chaining together operators (\texttt{Pipes}) using the Python Feature API (Section~\ref{sec:feature_api}).
\circled{2} \systemname parses the \texttt{Feature} into a dataflow graph.
\circled{3} \systemname's \texttt{Optimizer} (Section~\ref{sec:optimizer_static}) then collects performance statistics for each \texttt{Pipe} from past runs or via a short profiling job.
\circled{4} Using these statistics, the \texttt{Optimizer} applies optimizations via the Optimization interface (Section~\ref{sec:optimizer_api}) to yield an execution plan.
}

\rev{
\circled{5} Once training begins, \systemname uses the Execution interface (Section~\ref{sec:optimizer_api}) to 
initialize each \texttt{Pipe} by running its assigned \texttt{Variant} -- a \textit{physical} implementation of the operator -- on its respective engine.
\circled{6} Engines continuously process samples using their respective operators. %
\circled{7} Meanwhile, because training jobs ingest data in a data-parallel manner, \systemname creates a \texttt{Client} for each distributed training process (e.g., training node).
The \texttt{Client} launches \texttt{Driver} processes to manage operator processing across each engine and to ingest fully-processed samples into the ML training framework (e.g., PyTorch).
\circled{8} Finally, the \texttt{Client} continuously monitors processing and uses a \texttt{Scaler} (Section~\ref{sec:optimizer_dynamic}) to dynamically right-size the resources allocated to each engine via the Execution interface. 
}

\subsection{Feature API}\label{sec:feature_api}
\rev{
\systemname provides an easy-to-use \texttt{Feature} API.
It allows users to define dataflows (\texttt{Features}) by composing together stateless operators (\texttt{Pipes}) that implement common data loading primitives (e.g., batching, shuffling) and higher-order functions (e.g., map, filter) that support arbitrary Python UDFs.
}
Each \texttt{Pipe} applies a logical transformation to input samples, yielding transformed samples to the downstream \texttt{Pipe(s)}.
Transformations may be applied one-to-one (e.g., map), many-to-one (e.g., batch), or one-to-many (e.g., reading lines in a file).
\rev{
\texttt{Features} may also be non-linear DAGs with a single output node that generates mini-batches of data iteratively.
\texttt{Pipes} may ingest from/emit to one or more \texttt{Pipes} (e.g., zip/unzip).}

\begin{figure}[t]
  \centering
  \includegraphics[width=\linewidth]{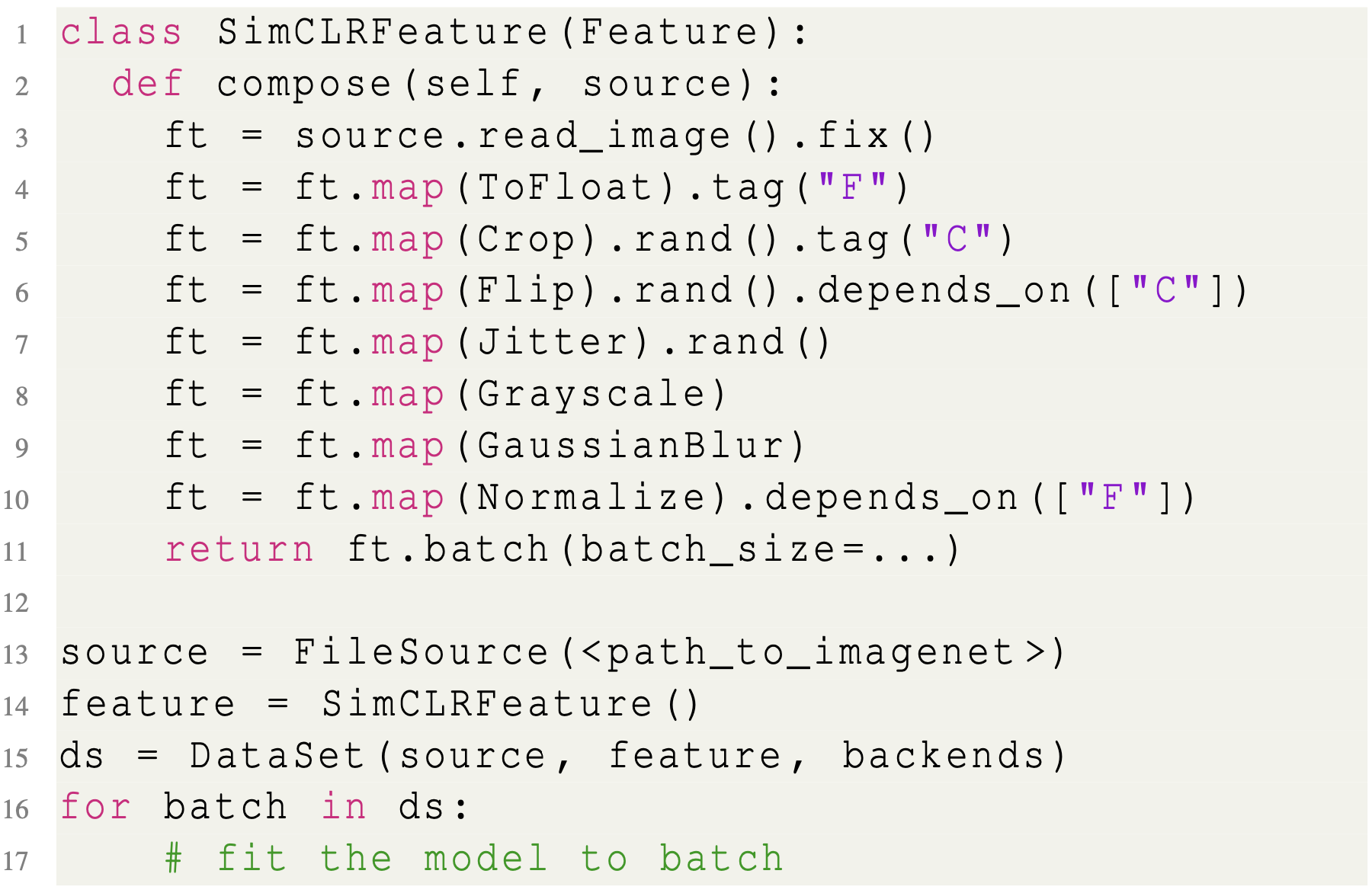} 
  \caption{\systemname \texttt{Feature} API.}
  \label{fig:feature_api}
\end{figure}

\rev{Users can thus easily define and share \texttt{Features}, without needing to manage their underlying execution details.}
For example, Figure~\ref{fig:feature_api} shows a \texttt{Feature} for the CV pipeline in Figure~\ref{fig:simclr}.
Each \texttt{Feature} is logical; it is not bound to a specific dataset, nor does it specify \textit{how} the dataflow is executed.
Instead, ML engineers define a \texttt{Source}, which wraps a raw dataset with a \texttt{Pipe}, providing an iterator over raw samples.
Users supply one or more \texttt{Sources}, the \texttt{Feature}, and available engines to \systemname, which constructs an iterable \texttt{DataSet}.
Training nodes simply iterate over the \texttt{DataSet}, which yields fully processed mini-batches within host memory to be consumed by ML frameworks.
Execution is lazy and incremental, allowing \systemname to scale to large out-of-memory datasets.

\noindent\textbf{Relaxed Operator Dependencies.}
While the original \texttt{Feature} dataflow specifies a \textit{potential} ordering, \systemname allows users to relax ordering constraints by expressing a dependency graph on top of the original dataflow specification.
As shown in Figure~\ref{fig:feature_api}, if a \texttt{Pipe} $b$ depends on $a$, users can label $a$ with an explicit $tag_a$ (Line 5) and declare the dependency via $b.depends\_on([tag_a])$ (Line 6).
Users can also fix the position of a pipe $a$ by calling $a.fix()$, which makes $a$ depend on all upstream \texttt{Pipes} and makes all downstream \texttt{Pipes} depend on $a$.
As we show in Section~\ref{sec:optimizer_static}, this allows the \texttt{Optimizer} to enumerate semantic-preserving reordered plans. 

\noindent\textbf{Random Operators.}
\systemname also allows ML engineers to designate which \texttt{Pipes} represent random augmentations of data (Lines 5-7), allowing the \texttt{Optimizer} to preserve randomness by disallowing caching after random operators. %

\noindent\textbf{Correctness and Fault Tolerance.}
\systemname ensures that the relaxed operator dependencies and randomness constraints are correctly met.
Specifically, if operator B depends on A, \systemname will never generate a plan where B precedes A.
Similarly, if operator C is random, \systemname will never insert a cache operator D such that C precedes D.
\rev{Users may choose to not specify dependencies or randomness; \systemname disables reordering and caching, respectively, to ensure correctness.
Inferring dependencies or randomness (e.g., via static code analysis~\cite{vldb5:hueske_blackbox, mlsys22:kuchnik_plumber}) is left for future work.
}

\systemname also guarantees precise checkpoints and exactly-once semantics, key requirements to ensure that model convergence is not affected by faults.
Specifically, each \texttt{Source} tags all training samples with a UUID, and the \texttt{Client} verifies receipt of all tags (accounting for aggregations and filters).
Because \texttt{Pipes} are stateless, upon detecting a fault, the \texttt{Client} instructs the \texttt{Source} to re-emit a specific sample to recompute the result.
\texttt{Clients} do not return duplicate samples to ensure exactly-once semantics.
\systemname provides a checkpoint API which allows \systemname to resume processing, skipping received samples, in the event of a training job or \texttt{Client} failure.

\begin{table}[t]
\centering
\caption{Optimization (top) \& Execution (bottom) interfaces.}\label{tbl:interface}
\small
\begin{tabular}{@{}ll@{}}
\toprule
API & Description \\ \midrule
register(pipe)           & Registers a new \texttt{Pipe} into the \texttt{DataSet}.                                             \\
fuse(pipes)              & Registers a new \texttt{Pipe} that fuses all input \texttt{Pipes}.              \\
update\_dfg(graph)                          & Updates the dataflow graph as specified.                     \\
assign(pipe, variant)                       & Assign a \texttt{Variant} to the provided \texttt{Pipe}.                                             \\
set\_shards(n)                      & Assign a number of \texttt{Drivers} for each \texttt{Client}.                              \\ \midrule
shard(n)                               & Shard the \texttt{Feature} into $n$ \texttt{Drivers}.                           \\
mutate(pipe, variant)   & Mutate the \texttt{Pipe} to the specified \texttt{Variant}.                              \\
scale(pipe, n) & Sets the parallelism of \texttt{Pipe} to $n$. \\ \bottomrule
\end{tabular}
\end{table}
\subsection{Optimization and Execution Interfaces}\label{sec:optimizer_api}
\noindent\textbf{Optimization Interface.}
The Optimization interface (Table~\ref{tbl:interface}) provides well-defined methods that allow the \texttt{Optimizer} to apply multiple optimizations to the \texttt{DataSet} in an extensible manner, \rev{allowing \systemname to easily integrate future optimizations}.
For example, to introduce caching or prefetching, the \texttt{Optimizer} creates a cache or prefetch \texttt{Pipe}, registers it via \texttt{register}, and updates the dataflow via \texttt{update\_dfg} to insert it at the appropriate location.
A fused \texttt{Pipe} can be created via \texttt{fuse} and inserted into the dataflow, and the dataflow can be reordered via \texttt{update\_dfg} accordingly.

\noindent\textbf{Execution Interface.}
\rev{The \texttt{Optimizer} also specifies pipeline execution via \texttt{Drivers} and \texttt{Variants}.
A \texttt{Driver} is an independent Python process that manages the end-to-end processing of a disjoint (data-parallel) subset of samples.
Each \texttt{Client} can use multiple \texttt{Drivers} to parallelize GIL-constrained Python operations within a training node.
Meanwhile, a \texttt{Variant} is a \textit{physical} implementation of the \texttt{Pipe}'s logical operation on a given engine (and potentially for a specific ML framework). %
For example, a \textit{map(tf.image.resize)} \texttt{Pipe} can have \texttt{Variants} that executes resize from the tf.image library a) locally in the \texttt{Driver}'s Python process, b) on a distributed worker, or even c) by using the runtime of another framework such as tf.data.
Native operators (e.g., batch) can further have specific \texttt{Variants} that process ML framework-specific tensors on a given engine (e.g., PyTorch/distributed or TF/local), which \systemname selects from based on the ML framework used by the training job.
}

Once the training job starts, each \texttt{Client} will create \texttt{Drivers} using \texttt{shard} and initiate processing for each \texttt{Pipe} by calling \texttt{mutate} to transform the \texttt{Pipe} to the appropriate \texttt{Variant}.
Some \texttt{Variants} allow a configurable amount of parallelism (e.g., process pool size); \texttt{scale} sets the amount of parallelism appropriately.
Throughout training, each \texttt{Client} ingests data and manages execution only for its respective training process (e.g., training node), allowing \systemname to support large-scale distributed training in a decentralized manner.
Specifically, each \texttt{Client} will use a local \texttt{Scaler} (Section~\ref{sec:optimizer_dynamic}) to tune both the parallelism and \texttt{Variant} of each \texttt{Pipe} (via \texttt{scale} and \texttt{mutate}) to meet its training process's throughput requirements. 

\section{Optimization and Dynamic Scaling}\label{sec:optimizer}
\systemname first applies global static optimizations to \rev{increase} the per-resource throughput.
During runtime, each \texttt{Client} then dynamically right-sizes the resources used by its \texttt{Variants} to efficiently meet the throughput demanded by its training process.

\noindent\textbf{Tracing and Profiling.}
The \texttt{Optimizer} relies on a collection of performance statistics within the \texttt{Metadata Store} (Figure~\ref{fig:block_diagram}) to calculate cost models across plans.
These statistics are automatically collected by \systemname, which traces the execution of each \texttt{Pipe}.
Each \texttt{Source} periodically marks \rev{emitted samples} to be traced.
Each \texttt{Pipe} transparently tags traced \rev{samples} with statistics, including the sample's execution latency and size (bytes), and the \texttt{Pipe}'s current \texttt{Variant} and prefetch buffer length (if applicable).
Upon reception, the \texttt{Client} updates the \texttt{Metadata Store} with traced results and a measure of the current overall throughput. %

The \texttt{Optimizer} requires a core set of statistics.
These include the throughput $tput_{base}$ of the baseline plan, $G_{base}$, which executes the un-optimized pipeline \textit{locally} (i.e., within a single \texttt{Driver} Python process). %
For each \texttt{Pipe} $p$, the \texttt{Optimizer} also requires the average latency to process a sample, $lat_{base}(p)$, and its average input and output sample sizes, $size_{in}(p)$ and $size_{out}(p)$, respectively.
Finally, for each \texttt{Variant} $v$ available for $p$, the \texttt{Optimizer} requires the average \texttt{DataSet} throughput achieved by offloading only $p$ to $v$, $tput_{v}(p)$.
If statistics are insufficient (e.g., from a previous job) \rev{or performance characteristics are different (e.g., the infrastructure changes)}, the \texttt{Optimizer} runs a short profiling job.

Profiling and optimization introduce negligible overheads (seconds) compared to long-running (hours-days) training jobs.

\subsection{Static Optimization}\label{sec:optimizer_static}
\rev{To explore the optimization search space, the \texttt{Optimizer} iteratively applies a set of cost- and rule-based optimization passes to the dataflow graph, similar to traditional database query optimizers~\cite{icde93:graefe_volcano, dataengbull95:graefe_cascades, jvldb22:kossman_datadependencies, vldb5:hueske_blackbox}.
Within each pass, the \texttt{Optimizer} begins by enumerating possible plans by applying a specific optimization technique to the current plan(s) (e.g., enumerating all possible operator orderings for reordering).
The \texttt{Optimizer} then evaluates the enumerated plans using a cost model or a set of rules.
To efficiently search the optimization space, each pass prunes plans based on cost and satisfiability (i.e., obeying user-specified dependencies and randomness constraints).
By default, each pass outputs the lowest-cost, permissible plan to the next pass.
Each optimization pass thus enumerates plans $\mathcal{G} = \{G_1, \cdots, G_n\}$ and determines the cost of a plan $G$ by calculating a $cost(p)$ for each \texttt{Pipe} $p \in G$.
It aims to find:}
\begin{equation}
    \rev{G^* = \operatorname*{argmin}_{G \in \mathcal{G}} \sum_{p \in G} cost(p), \quad \text{s.t. } G \text{ satisfies user constraints}}
\end{equation}
\rev{A higher cost represents more work and thus lower performance.
\systemname uses a comprehensive cost model that extends $cost(p)$ with each pass, allowing system experts to customize the cost model used by a pass, if needed, in a modular manner.}

\rev{The \texttt{Optimizer} uses an initial cost model for the profiled baseline plan $G_{base}$, which weights the cost of each pipe by its fractional latency in the end-to-end pipeline:}
\begin{equation}
    \rev{cost_{base}(p) = \frac{lat_{base}(p)}{\Sigma_{i \in G_{base}} lat_{base}(i)} / tput_{base}}
\end{equation}

\rev{We next describe each optimization pass in the order that it is applied.
We applied logical passes (e.g., reordering) prior to physical optimizations (e.g., offloading).
Within each optimization pass, we present a) how the \texttt{Optimizer} enumerates plans, and b) (if applicable) the cost model used.
Using iterative passes allows the \texttt{Optimizer} to be easily extended with further optimizations.}

\noindent\textbf{Reordering.}
The \texttt{Optimizer} first finds the best dataflow ordering.

\rev{\textit{Enumeration.}}
The \texttt{Optimizer} enumerates all permissible reorderings of the initial plan $G_{base}$.
It does so by removing the output \texttt{Pipe}, $o$, of $G_{base}$ and recursively calculating the set of all possible reorderings of the shrunk graph.
For each shrunk reordering $G_s$ with output \texttt{Pipe} $o_s$, adding $o$ as a successor to $o_s$ produces a viable reordering.
Furthermore, if $o$ and $o_s$ may be reordered according to user constraints, swapping $o$ and $o_s$ also produces a viable reordering (i.e., $o$ precedes $o_s$).
\texttt{reorder} only reorders \texttt{Pipes} within their linear subgraph (i.e., \texttt{fix()}-ing \texttt{Pipes} with $>1$ input or output).

\rev{\textit{Cost model.}}
To calculate the cost of a reordering $R$, reordering augments the base cost model by calculating a \textit{size scaling factor} $S(p) = size_{out,base}(p) / size_{in,base}(p)$ for each pipe $p$, representing how it scales the size of its output on average.
The cost model then computes the new input size of $p \in R$ as $size_{in, R}(p) = size_{raw} * \Pi_{i \in U} S(i)$, where $size_{raw}$ is the average raw sample size and $U \subset R$ is the ordered sequence of all ancestors of $p$.
The cost model calculates the reordered cost as:
\begin{equation}\label{eq:cost_reordering}
    \rev{cost_{R}(p) = (size_{in, R}(p) / size_{in, base}(p)) * cost_{base}(p)}
\end{equation}
\rev{Thus, reordering assumes that the cost of each \texttt{Pipe} scales linearly with its input sample size (e.g., a tokenizer requires half as much compute to process half as many tokens).
However, system experts may customize the cost model for different scaling properties of each pipe (e.g., to scale $cost_{base}(p)$ quadratically with sample size).}
Furthermore, since $size_{in}$ and $size_{out}$ is the average sample size, reordering optimizes the location of operators that change both selectivity (e.g., filter) and sample size (e.g., crop).

\noindent\textbf{Caching.}
Next, the \texttt{Optimizer} evaluates if and where to best cache (i.e., materialize) intermediate data.

\rev{\textit{Enumeration.}
The \texttt{Optimizer} enumerates plans by creating a new plan for each permissible caching location within the dataflow (i.e., after every operator that does not contain an ancestor that is marked random).
We currently consider only inserting one cache operator in the dataflow (i.e., finding the best cache location).
}

\rev{\textit{Cost model.}}
To calculate the cost of a plan with a cache \texttt{Pipe} $p_{cache}$, the cost model simply sets the cost of any exclusive ancestor $p$ of $p_{cache}$ (i.e., all paths from $p$ to the output contain $p_{cache}$) to zero.
To account for I/O costs to read cached data, we calculate
\begin{equation}
    \rev{cost(p_{cache}) = d * size_{in, R}(p_{cache})}
\end{equation}
where $d$ is a constant derived from the node's disk I/O throughput, and $size_{in, R}(p_{cache})$ is derived using $S(p)$ as with reordering.

\noindent\textbf{Fusion and Offloading.}
The \texttt{Optimizer} considers offloading and fusion concurrently.

\rev{\textit{Enumeration.}}
It enumerates all possible offloading plans by generating a set $P_i = \{(p_i, v)| v \in V$ and $p_i \text{ supports } v\}$ for each \texttt{Pipe} $p_i$, which contains the set of supported \texttt{Variants} for $p_i$ within the user-provided engines $V$.
The \texttt{Optimizer} first enumerates plans by taking the Cartesian product between all $P_i$s.
Then, the \texttt{Optimizer} performs all possible fusions for each plan (e.g., fusing adjacent \texttt{Pipes} assigned the same \texttt{Variant} if each \texttt{Pipe} supports fusion).
\rev{Furthermore, as mentioned in Section~\ref{sec:optimizations_requirements}, caching may preclude the ability to fuse operators.
The \texttt{Optimizer} thus enumerates plans based on the input plan with and without the inserted cache \texttt{Pipe}.
}

\rev{\textit{Cost model.}
To compare costs between plans, the cost model uses Amdahl's Law to determine the benefit that $p$ gains if it is offloaded to a \texttt{Variant} $v$.}
It assigns a lower cost $cost_{v}(p)$ to \texttt{Variants} that achieve a higher throughput.
The cost model uses the overall speedup $s_v(p) = tput_v(p) / tput_{base}$ of $v$.
It then solves for the speedup of $p$ on $v$ and linearly scales the reordered $cost_{R}(p)$ accordingly.
We constrain $cost_{v}(p) \geq 0$.
\begin{equation}
    cost_{v}(p) = cost_{R}(p)*\frac{s_v(p)^{-1} - (1 - f(p))}{f(p)}
\end{equation}

Fusion reduces I/O costs between \texttt{Pipes}.
To calculate the cost of a fused \texttt{Pipe} $p$, which fuses pipes $q_1, ..., q_n$, the cost model calculates the reduction of I/O relative to the un-fused baseline as $io(p) = (size_{in, R}(q_1) + size_{out, R}(q_{n})) / (size_{in, R}(q_1) + 2*size_{in, R}(q_2) + ... + 2*size_{in, R}(q_n) + size_{out, R}(q_{n}))$.
The cost model then discounts the aggregate costs of all original \texttt{Pipes} by the relative I/O savings:
\begin{equation}
\rev{cost_{fused, v}(p) = io(p) * (cost_{v}(q_1) + \cdots + cost_{v}(q_n))}
\end{equation}

\noindent\textbf{Prefetching and Sharding.}
Finally, the \texttt{Optimizer} applies a set of rules to prefetch and shard the \texttt{DataSet}.
It inserts a prefetching \texttt{Pipe} after each offloaded (non-\textit{base}) \texttt{Variant}, as well as at the end of the dataflow, to allow pipelined execution throughout the input data pipeline.
The \texttt{Optimizer} also calculates the ideal number of shards (i.e., \texttt{Drivers}) to use for each \texttt{Client}.
To do so, it estimates the throughput of each \texttt{Driver} using the cost model.
If the throughput is over a threshold, the \texttt{Optimizer} runs a single \texttt{Driver} \textit{within} each \texttt{Client} process to avoid introducing an inter-process communication bottleneck.
Otherwise, the \texttt{Optimizer} further replicates \texttt{Drivers} to \rev{improve} local parallelism (i.e., one per host CPU core).

\rev{
\noindent\textbf{Summary.}
The \texttt{Optimizer} applies complex optimizations by enumerating and pruning plans in discrete passes (like query optimizers), allowing \systemname to search the optimization space in an efficient and extensible manner.
Each cost-based optimization pass leverages an informed cost model, based on
well-founded principles (e.g., Amdahl's Law), that estimates the cost of each \texttt{Pipe} using profiled statistics.
Some models apply heuristics to support black-box UDFs; for example, reordering assumes that operator costs scale linearly with input size.
However, users may easily customize the model in a modular manner for each \texttt{Pipe} if needed, for example by modifying Equation~\ref{eq:cost_reordering} to scale quadratically w.r.t. input size.
}

\subsection{Dynamic Scaling}\label{sec:optimizer_dynamic}
The \texttt{Optimizer}'s static pass is designed to \rev{select a high-throughput plan in order to increase the utilization of training accelerators.} %
Since \texttt{Clients} independently process samples for its respective training process, each \texttt{Client} runs an \rev{\texttt{Scaler}} during training to \rev{right-size resources to efficiently} meet its training process's throughput demands.
The \rev{\texttt{Scaler}} continuously monitors performance, identifies the bottleneck \texttt{Pipe}, and tunes its \rev{parallelism}. %

As mentioned above, each \texttt{Client} continuously traces and reports runtime metrics during training.
To identify the bottleneck, the \rev{\texttt{Scaler}} examines the prefetch buffer at the pipeline output. %
If the buffer length is over a configurable threshold, the input data pipeline is not the bottleneck.
In this case, the \texttt{Scaler} selects a random \texttt{Pipe} $p$ with a non-\textit{base} \texttt{Variant} and scales down its parallelism by a unit (e.g., one process).
Importantly, if the current parallelism of $p$ cannot be further decreased, \systemname will mutate $p$ into the \textit{base} \texttt{Variant} to avoid over-provisioning resources. %

However, if the output buffer is below a threshold, a bottleneck exists.
The \rev{\texttt{Scaler}} will attempt to scale-up the parallelism for the bottleneck \texttt{Pipe} $p_b$.
It identifies $p_b$ by examining the set of all \texttt{Pipes} which were statically assigned a non-\textit{base} \texttt{Variant}, $P^*$, which represents \texttt{Pipes} that benefit from offloading.
First, the \rev{\texttt{Scaler}} examines the prefetch buffer of all $p \in P^*$ that are currently offloaded (i.e., non-\textit{base}), selecting the $p$ with the smallest buffer below a threshold.
If no such \texttt{Pipes} exist (e.g., if all $p \in P^*$ are mutated to the \textit{base} \texttt{Variant}), the \rev{\texttt{Scaler}} examines all $p \in P^*$ with \textit{base} \texttt{Variants} and selects the $p$ with the largest $lat_{base}(p)$ -- the largest speedup opportunity.
\shep{
Given $p_b$, the \texttt{Scaler} will iteratively increase its parallelism by a unit until throughput plateaus, potentially mutating $p_b$ back into a non-\textit{base} \texttt{Variant}.
If the backend's resources are exhausted, the \texttt{Scaler} will scale down another random \texttt{Pipe} with the same \texttt{Variant} as $p_b$.
The \texttt{Scaler} periodically runs (e.g., every minute), scaling resources to meet throughput demands.
}

\shep{The \texttt{Optimizer}'s static passes (Section~\ref{sec:optimizer_static}) are responsible for exploring the complex optimization space presented in Section~\ref{sec:optimizations_space}.
Meanwhile, the \texttt{Scaler} is responsible for scaling the resources used to execute the optimized execution plan in order to meet throughput demands.
While alternative methods such as Bayesian Optimization~\cite{cal21:li_rambo} or simulated annealing~\cite{ipdps04:juedes_heuristic} could be applied to scaling, we find that the \texttt{Scaler}'s hill-climbing approach is a simple and effective solution.
This is because as we increase the parallelism (and thus throughput) of a \texttt{Pipe}, we can apply Amdahl's Law~\cite{afips67:amdahl} to model the throughput of the entire pipeline (as in Section~\ref{sec:optimizer_static}), resulting in a concave function with respect to parallelism.
A similar concave speedup function has long been used to scale parallelism for cases ranging from multicore processors~\cite{computer08:hill_amdahl, sigmetrics09:yao_extending-amdahl} to datacenter workloads~\cite{hpca18:zahedi_amdahl, thesis:berg}.
We evaluate the \texttt{Scaler} in Section~\ref{sec:eval_scaling_caching}, and Figure~\ref{fig:scaling} experimentally validates this model.}

\section{Evaluation}\label{sec:evaluation}
We designed \systemname to support the numerous libraries and frameworks currently used across ML deployments. %
Since ML practitioners predominantly rely on Python, \systemname is built from the ground up in $\sim$12K lines of Python.
\systemname supports all popular ML frameworks (e.g., PyTorch, JAX, and TensorFlow) and can execute operations from arbitrary preprocessing libraries that provide a Python API. 

\noindent\textbf{Workloads.}
\begin{table}[t]
\centering
\caption{Description of pipelines used to evaluate \systemname.}\label{tbl:pipelines_eval}
\small
\begin{tabular}{@{}c|c|c@{}}
\toprule
Pipeline  & Description & \rev{Model} \\ \midrule
\thead{\rev{CV-}\\\rev{\{torch,tf\}}}  & \thead{Decode $\rightarrow$ Float $\rightarrow$ RandCrop $\rightarrow$ RandFlip \\ $\rightarrow$ Jitter $\rightarrow$ Grayscale $\rightarrow$ Blur $\rightarrow$ Normalize}   & \thead{\rev{SimCLR}\\ \rev{\cite{arxiv:chen_simclr, arxiv:chen_simclrv2}}} \\ \hline

\thead{\rev{SSD}-\\\rev{\{torch,tf\}}} & \thead{\rev{Decode $\rightarrow$ Resized Bounding Box Crop $\rightarrow$} \\ \rev{Flip $\rightarrow$ Distort $\rightarrow$ Normalize}} & \thead{\rev{SSD}\\ \rev{\cite{arxiv:liu_ssd}}} \\ \hline

\thead{\rev{NLP}-\\\rev{\{torch,hf-tf,tf\}}} & \thead{Read $\rightarrow$ Tokenize $\rightarrow$ Truncate $\rightarrow$ Embedding}   & \thead{\rev{LSTM}\\ \rev{\cite{neco97:hochreiter_lstm}}}     \\ \hline

\thead{ASR}       & \thead{Decode $\rightarrow$ Resample $\rightarrow$ Spectrogram $\rightarrow$ \\ Stretch $\rightarrow$ Time Mask $\rightarrow$ Freq. Mask $\rightarrow$ Mel Scale}  & \thead{\rev{RNN-T} \\ \rev{\cite{arxiv:graves_rnnt}}}  \\ \bottomrule

\end{tabular}
\end{table}
We evaluated \systemname on a diverse set of \rev{eight} ML input data pipelines across computer vision, natural language, and speech domains as shown in Table~\ref{tbl:pipelines_eval}.
\rev{\texttt{CV-torch} and \texttt{CV-tf} implemented the SimCLR~\cite{arxiv:chen_simclr, arxiv:chen_simclrv2} pipeline in Figure~\ref{fig:simclr} using PyTorch and TensorFlow (TF) operators, respectively.
They processed the ImageNet dataset~\cite{cvpr09:deng_imagenet} and used the semantic constraints shown in Figure~\ref{fig:feature_api}.
\texttt{SSD-torch} and \texttt{SSD-tf} implemented the SSD pipeline from the MLPerf Training benchmark~\cite{arxiv:mlperf_training} using PyTorch and TF operators, respectively.
\texttt{SSD} pipelines used the COCO dataset~\cite{arxiv:lin_coco}.
All \texttt{SSD} operators were random; Distort was able to be reordered between Decode and Normalize.
\texttt{NLP-torch}, \texttt{NLP-hf}, and \texttt{NLP-tf} implemented a standard pipeline for natural language tasks~\cite{techreport:radford_gpt2, arxiv:gpt3}.
\texttt{NLP-torch} used torchtext~\cite{website:torchtext} operators, while \texttt{NLP-hf-tf} and \texttt{NLP-tf} used tf.text~\cite{website:tftext} operators with a Hugging Face~\cite{website:huggingface-tokenizers} and TF tokenizer, respectively.
All \texttt{NLP} pipelines used the WikiText-103 dataset~\cite{arxiv:merity_wikitext}, and all operators were not random and not reorderable.
Finally, \texttt{ASR} implemented the SpecAugment speech recognition pipeline~\cite{interspeech19:park_specaugment} using the third-party librosa~\cite{scipy15:mcfee_librosa} library.
\texttt{ASR} used the Common Voice dataset~\cite{arxiv:ardila_commonvoice}.
All \texttt{ASR} operators were fixed except for Stretch, TimeMask, and FreqMask; TimeMask and FreqMask were random.
}

\begin{figure*}[t!]
  \centering
  \includegraphics[width=\textwidth]{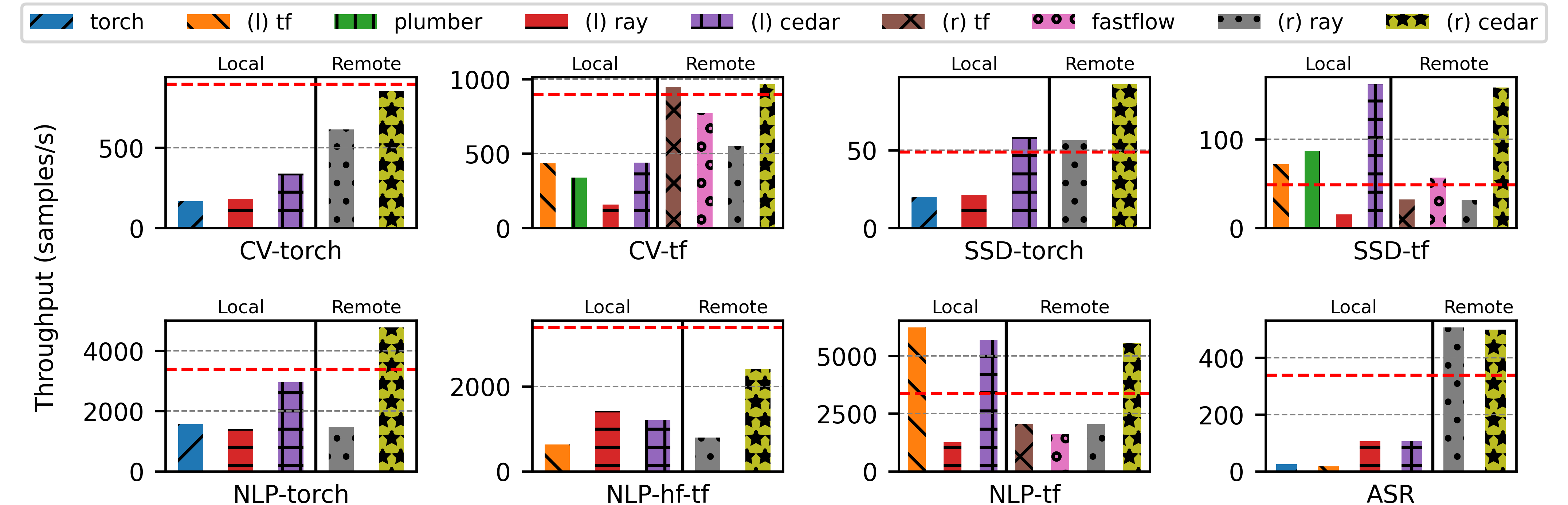} 
  \caption{\rev{Achieved processing throughput ($T_p$, higher is better) across eight pipelines. (l) and (r) denote the \texttt{local} and \texttt{remote} setups of each system, where applicable. The dashed red line marks the GPU computation rate ($T_g$) for the corresponding model of each pipeline shown in Table~\ref{tbl:pipelines_eval} on an NVIDIA A100 GPU. Incompatible system/pipelines pairs are not shown.}}
  \label{fig:aggregate_eval}
\end{figure*}

\subsection{\systemname Optimizer}
\rev{
We begin by evaluating how effectively \systemname's static optimizations improve per-resource throughput compared to state-of-the-art input data systems.
As highlighted by recent work~\cite{sigmod22:isenko_presto, vldb14:mohan_coordl}, end-to-end training throughput $T_{e2e}$ is the minimum of the input data throughput $T_p$ and the GPU computation rate $T_g$.
$T_g$ depends on the model and training infrastructure (e.g., GPU version and networking hardware).
The primary goal of an input data optimizer is to increase per-resource throughput (e.g., $T_p$ per CPU core).
If $T_p < T_g$ given fixed input data resources (e.g., the training node's host CPU), increasing the per-resource throughput, and thus $T_p$, directly improves $T_{e2e}$.
Alternatively, if $T_p > T_g$, increased per-resource throughput allows a dynamic scaler (which we evaluate in Section~\ref{sec:eval_scaling_caching}) to reduce the amount of allocated resources, improving the resource efficiency of input data systems.
}

\rev{
Thus, to directly evaluate the per-resource throughput, we evaluated the maximum $T_p$ achieved by systems on two hardware setups.
Since training jobs often use CPU host resources for input data processing, we first used a \texttt{local} setup where all systems performed processing on a single 8-core VM (\texttt{n2-standard-8} on Google Cloud).
We provided \systemname with a Python multiprocessing engine, as well as two engines that used a tf.data or a Ray Data runtime to execute operators, respectively.
In this setup, we compared against tf.data~\cite{vldb14:murray_tfdata}, Plumber~\cite{mlsys22:kuchnik_plumber}, Ray Data~\cite{website:ray-data}, and the PyTorch DataLoader~\cite{website:torch-utilsdata}.
We also used a distributed (\texttt{remote}) setup, which provided each system with a remote 32-core VM (\texttt{n2-standard-32}) in addition to the \texttt{local} VM.
In addition to the \texttt{local} engines, we provided \systemname with a \texttt{distributed} engine which executed operators on the remote VM.
In the \texttt{remote} setup, we compared against tf.data service~\cite{socc23:audibert_tfdataservice}, FastFlow~\cite{vldb16:um_fastflow}, and Ray Data~\cite{website:ray-data}.
}

\rev{
We configured each system to maximize throughput (e.g., enabled auto-tuning in tf.data). 
We also disabled caching in order to directly evaluate the improvements to computational efficiency (e.g., avoiding degenerate cases where the pipeline output is cached).
We explicitly evaluate caching in Section~\ref{sec:eval_scaling_caching}.
Figure~\ref{fig:aggregate_eval} shows the input data throughput ($T_p$) for each system across the eight pipelines.
To highlight the importance of input data optimizations on $T_{e2e}$, we also report the $T_g$ of the representative model for each pipeline shown in Table~\ref{tbl:pipelines_eval} on an A100 GPU.
Because input data pipelines are data-parallel (i.e., independent across training processes), these results directly scale to multi-GPU training environments as each GPU would require a similar input data demand.
}

\noindent\rev{\textbf{\systemname optimizes local input data processing throughput.}
For the \texttt{CV} and \texttt{SSD} pipelines, \systemname reordered operators to reduce the overall amount of computation required per sample, which we further explore in Section~\ref{sec:eval_ablation}. This allowed \systemname to largely out-perform PyTorch, tf.data, Plumber, and Ray Data.
For \texttt{CV-tf}, tf.data and Plumber were able to achieve a comparable performance due to their underlying TF graph optimizer and performant C++ backend.
}

\rev{
For \texttt{NLP-torch}, both PyTorch and Ray Data suffered from a serialization bottleneck in sending large embeddings between processes.
\systemname was able to avoid this by fusing Read, Tokenize, and Truncate into one multiprocess \texttt{Pipe}, while performing Embedding within the main \texttt{Client} process.
For \texttt{NLP-hf-tf}, tf.data could not optimize the non-TF Tokenizer, while Ray Data achieved slightly higher performance ($1.17\times$) than \systemname because its Arrow core eliminated intermediate data copies.
\systemname's extensibility allows it to adopt Arrow as a future \texttt{Variant}.
For \texttt{NLP-tf}, tf.data was able to compile the pipeline into an optimized TF graph.
\systemname recognized this benefit and offloaded execution to the tf.data engine, \textit{without requiring users to modify the pipeline}, allowing \systemname to near tf.data's throughput ($0.91\times$ due to tracing) and out-perform Ray Data.
}

\rev{
For \texttt{ASR}, \systemname first reordered Stretch (to decrease work for the Mask operators) and then offloaded execution to the Ray Data engine, obtaining its zero-copy benefits and matching Ray Data's performance.
Meanwhile, PyTorch and tf.data suffered from copy overheads, and tf.data could not optimize the non-TF operators.
}

\rev{
These results highlight the impact of \systemname's extensible optimizer -- optimizing the dataflow via reordering and fusion, and offloading and prefetching execution to the best engine -- all without user input.
\systemname out-performed tf.data, Plumber, Ray Data, and PyTorch by up to $6.14\times$, $1.87\times$, $10.65\times$, and $4.28\times$ on the \texttt{local} setup, respectively.
In almost all cases, the input data throughput $T_p$ achieved by baselines was less than the GPU rate $T_g$.
Since $T_{e2e}=min(T_p, T_g)$, \systemname's ability to improve $T_p$ over the baselines directly translates to a corresponding improvement in end-to-end training performance.}

\noindent\rev{\textbf{\systemname intelligently uses distributed processing engines.}
For the \texttt{CV} and \texttt{SSD} pipelines, \systemname generated a similar reordering the \texttt{local} setup.
\systemname also offloaded compute-intensive operators (e.g., Distort for \texttt{SSD} and Jitter/Blur for \texttt{CV}) to the \texttt{distributed} engine for \texttt{SSD-torch} and the \texttt{CV} pipelines, further improving its throughput over the \texttt{local} case.
Interestingly, \systemname did not use the \texttt{distributed} engine for \texttt{SSD-tf}.
It instead correctly recognized that offloading operators remotely would incur a slowdown due to data movement overheads.
\systemname improved throughput over tf.data service, FastFlow, and Ray Data; tf.data service matched \systemname's performance for \texttt{CV-tf} due to its TF graph optimizations.
}

\rev{
For \texttt{NLP-torch} and \texttt{NLP-hf-tf}, \systemname fused and offloaded only the Read, Tokenize, and Truncate operators to the \texttt{distributed} engine, avoiding embedding serialization overheads as in the \texttt{local} case.
\systemname correctly decided to not offload any \texttt{NLP-tf} operators to the remote VM, avoiding network overheads similar to \texttt{SSD-tf}.
Meanwhile, tf.data service and Ray Data were both limited by these communication overheads.
FastFlow executed processing locally, but inserted logic that hampered TensorFlow's graph compiler.
Finally, \systemname automatically determined that reordering and offloading \texttt{ASR} to a distributed Ray Data engine was ideal for the same zero-copy benefits as the \texttt{local} setup, matching Ray Data.
}

\rev{
\systemname was able to effectively and judiciously leverage distributed processing engines, out-performing Ray Data, FastFlow, and tf.data service by up to $4.99\times$, $3.45\times$, and $4.94\times$, respectively.
By improving the achievable $T_p$ compared to baselines, given the same remote VM, \systemname reduces the substantial input data resources needed to meet GPU demands.
For instance, linearly scaling the Ray Data's $T_p$ per core to match the $T_g$ demand for \texttt{CV-torch} would require remote CPU cores equal in cost to \textit{$83\%$ of the A100 GPU VM} itself (even discounting the cost of the local VM), based on Google Cloud billing rates.
Meanwhile, \systemname reduces this cost to $56\%$; the \texttt{Scaler} (which we evaluate next) can use fewer resources to match $T_g$ demands.
}

\rev{
Finally, \textbf{\systemname supports and optimizes diverse input data pipelines.}
In contrast, many input data systems could not support all of the evaluation pipelines.
DataLoaders and tf.data were limited to PyTorch and TF pipelines, respectively.
Systems reliant on TF graphs -- tf.data service, Plumber, and FastFlow -- could not use any non-TF operator, such as a Hugging Face tokenizer or librosa.
Plumber could also not support operators that required non-serializable assets, such as the TF tokenizer in \texttt{NLP-tf}.
}

\begin{figure}[t]
  \centering
  \includegraphics[width=\linewidth]{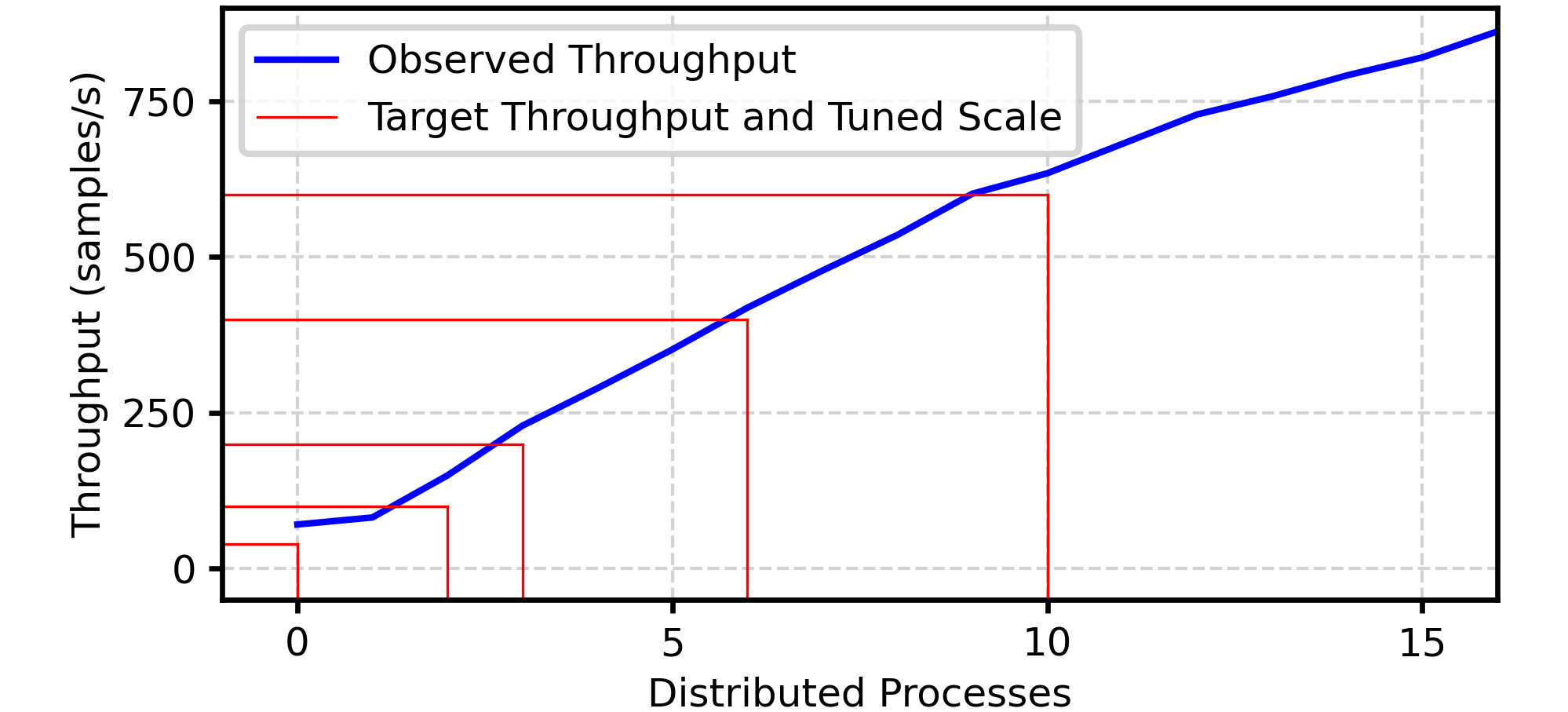} 
  \caption{\rev{Throughput as the number of distributed processes increases for the \texttt{CV-torch} pipeline. The red box shows the scale found by the \texttt{Scaler} given a target throughput.}}
  \label{fig:scaling}
\end{figure}

\subsection{Dynamic Scaling and Caching}\label{sec:eval_scaling_caching}
\rev{
\noindent\textbf{\systemname adjusts parallelism to efficiently meet diverse training throughput demands.}
While \systemname's \texttt{Optimizer} successfully improves per-resource throughput, its \texttt{Scaler} is responsible for translating this high performance to high resource efficiency by right-sizing resources to meet a given throughput demand (i.e., matching $T_p$ to $T_g$).
To evaluate this, we used the \texttt{CV-torch} pipeline with the \texttt{remote} setup (other pipelines showed similar results), which \systemname optimized by fusing and offloading compute-intensive operators (Jitter and Blur) to the \texttt{distributed} engine.
}

\rev{
The blue line in Figure~\ref{fig:scaling} shows the throughput (i.e., $T_p$) achieved by \systemname as we swept the number of \texttt{distributed} processes, up to saturation.
We next set 5 target training throughputs (i.e., $T_g$), shown by the horizontal red lines.
We allowed the \texttt{Scaler} to adjust the number of processes; the tuned scale for each target is shown by the intersecting vertical red line.
An efficient input data system should \textit{tune parallelism such that this intersection is close to, but below the blue line}.
This means that the system provisions the minimal amount of resources to meet demand (i.e., $T_p > T_g$).
}

\rev{
Figure~\ref{fig:scaling} shows that \systemname can not only scale across a wide range of throughput demands, but also efficiently right-size resources; it selected the smallest amount of parallelism to meet each target.
Furthermore, \systemname's ability to \textit{dynamically} mutate a \texttt{Pipe}'s \texttt{Variant} even completely deallocated the remote VM in the case of low training throughput, as shown by the zero processes result in Figure~\ref{fig:scaling}.
This is in contrast to systems such as tf.data service~\cite{socc23:audibert_tfdataservice} and Cachew~\cite{atc22:graur_cachew} which must \textit{always} use distributed processing.
}

\begin{table}[t]
\centering
\caption{Auto-cached throughput (normalized to \texttt{remote} \systemname, higher is better) and throughput of the next best cache location across three torch pipelines.}\label{tbl:caching_eval}
\small
\begin{tabular}{@{}l|ll@{}}
\toprule
Pipeline  & Norm. Throughput with Caching & Next Best Throughput\\ \midrule
CV-torch  & 1.74                        & 1.23                    \\
NLP-torch & 1.42                                 & 1.37                    \\
ASR       & 1.00 (Do Not Cache)                    & 0.82                    \\ \bottomrule
\end{tabular}
\end{table}
\noindent\textbf{\rev{\systemname optimizes \textit{if} and \textit{where} to apply caching.}}
To evaluate \systemname's ability to leverage caching, we allowed \systemname to automatically place a cache \texttt{Pipe}, which materialized all intermediate samples to disk in the first epoch.
We then ran multiple epochs of the torch pipelines across three domains (\texttt{CV-torch}, \texttt{NLP-torch}, and \texttt{ASR}); \rev{caching is not applicable to \texttt{SSD} as it only used random operators}.
Table~\ref{tbl:caching_eval} shows the throughput (higher is better) of the cached plan after the first epoch, normalized to the throughput achieved by the plan generated by \texttt{cedar-remote}.
We also report the \textit{next-best} throughput achieved by enumerating all other cache locations.
\systemname was able to find the optimal location to apply caching.

For \texttt{CV-torch}, \systemname cached the result after Decode and Grayscale, but prior to random Cropping, satisfying randomness requirements.
This improved throughput by reducing disk I/O (Decode) and compute (Grayscale).
For \texttt{NLP-torch}, \systemname cached the tokenized and truncated sample \textit{prior} to Embedding, avoiding overheads of reading large embeddings.
Interestingly, \systemname did \textit{not} cache \texttt{ASR}, determining that re-computation was ideal because transforms significantly increased data volumes.
Table~\ref{tbl:caching_eval} confirms this; the next ``best'' solution cached the output of the entire pipeline, which \textit{reduced} throughput by $18\%$.
\systemname is effectively able to apply caching while reasoning about its complex interactions with other optimizations. 

\subsection{In-depth Analysis}\label{sec:eval_ablation}
\begin{figure}[t]
  \centering
  \includegraphics[width=\linewidth]{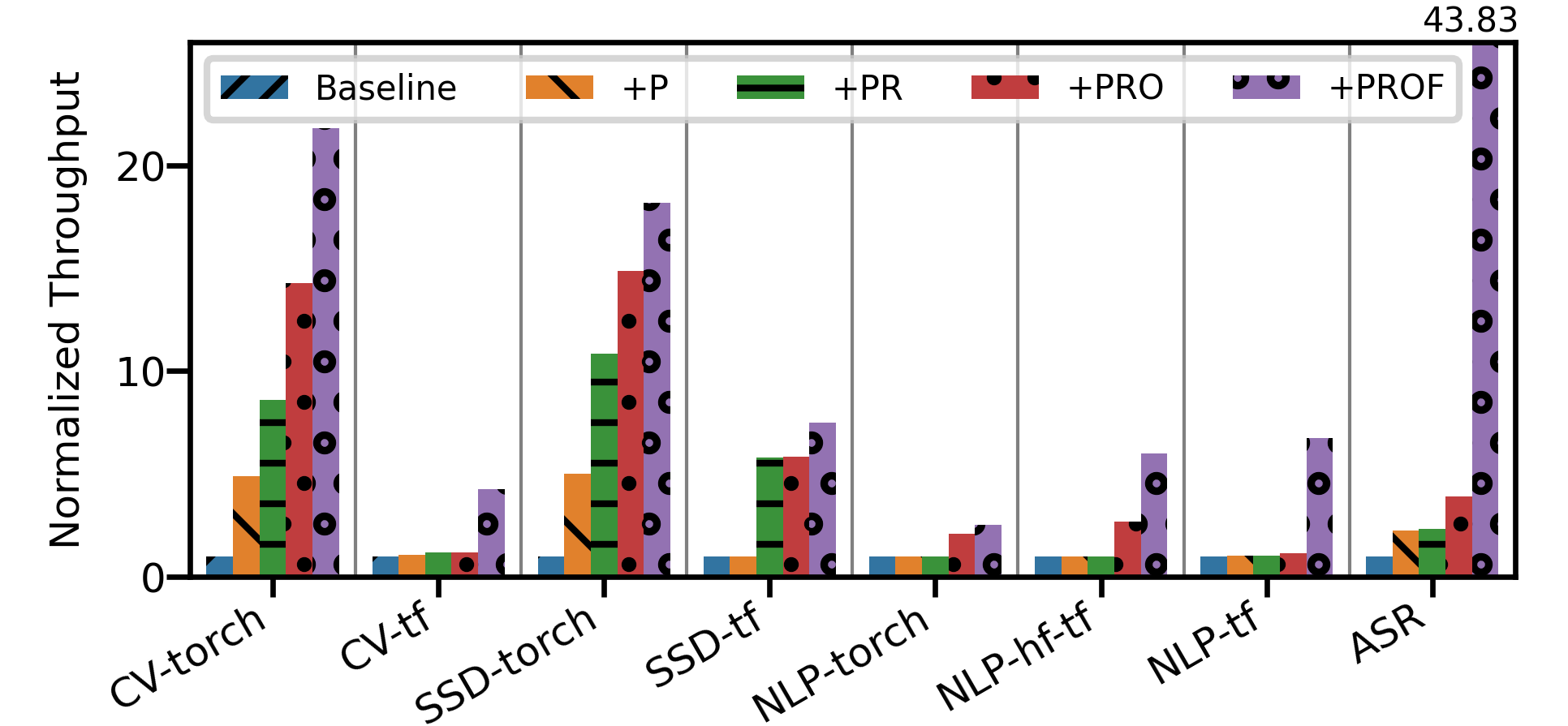} 
  \caption{\rev{Ablation study showing throughput across pipelines as optimization techniques are successively enabled, normalized to the unoptimized pipeline. P = local parallelism, R = reordering, O = offloading, F = fusion.}}
  \label{fig:ablation_eval}
\end{figure}

\noindent\textbf{Combining optimizations is essential to \systemname's performance.}
To understand \systemname's ability to \textit{combine} optimizations, we performed an ablation study by successively enabling local parallelism (i.e., multiple \texttt{Drivers}), reordering, offloading (enabling the \texttt{distributed} engine), and fusion.
All experiments used prefetching.
Figure~\ref{fig:ablation_eval} shows the \rev{throughput} of each experiment, normalized to the baseline (i.e., executing the unoptimized \texttt{DataSet} within a single \texttt{Driver}).
No single optimization is a panacea. %
Instead, a diverse set of optimizations is needed to achieve high performance due to the diverse characteristics across pipelines.

Local parallelism was effective for \texttt{CV-torch}, \rev{\texttt{SSD-torch}}, and \texttt{ASR} since using multiple \texttt{Drivers} bypassed their GIL bottleneck.
\texttt{CV-tf} and \rev{\texttt{SSD-tf}} largely used multithreaded C++ operators, limiting this benefit. 
\systemname did not parallelize the NLP pipelines to avoid serialization overheads.
As we explore next, reordering was effective at improving both \texttt{CV} and \texttt{SSD} pipelines since they used size-changing operators.
Meanwhile, \texttt{ASR} had more limited reordering opportunities, and the \texttt{NLP} pipelines were not able to be reordered.
Most pipelines took advantage of offloading across engines, but the overheads incurred by data movement across offloaded pipes limited its effectiveness for some pipelines.
By eliminating these overheads with fusion, \systemname was ultimately able to improve throughput by $2.54-43.82\times$ compared to the baseline.
The successive improvements with each step showcase \systemname's ability to systematically apply and combine optimizations.

\noindent\rev{\textbf{Reordering improves throughput by eliminating wasted work on a per-sample basis.}
While reordering can reduce wasted work based on operator selectivity (see Section~\ref{sec:optimizer_static}), akin to predicate pushdown in traditional query optimizers, \systemname further extends this benefit to operators that affect the size of individual samples.
The impact of this is shown in Figure~\ref{fig:ablation_eval}; reordering increased throughput by up to $5.79\times$ over parallelism alone.
Specifically, for the \texttt{CV} pipelines, \systemname reordered operators that reduced the size of each image (Crop, Grayscale) towards the beginning of the pipeline, and size-increasing operators (Float) towards the end.
Meanwhile for the \texttt{SSD} pipelines, \systemname moved Distort before Resized Crop, as the Resized Crop increased image sizes.
This significantly reduced the necessary compute for each sample while obeying the dependency constraints of each pipeline.
}

\begin{table}[t]
\centering
\caption{\rev{Throughput overheads with tagging/tracing enabled.}}\label{tbl:tracing}
\small
\begin{tabular}{@{}lllll@{}}
\toprule
         & CV-torch & SSD-torch & NLP-torch & ASR    \\ \midrule
Throughput Loss & 1.45\%   & 2.08\%    & 7.12\%    & 0.71\% \\ \bottomrule
\end{tabular}
\end{table}
\noindent\rev{\textbf{\systemname introduces minimal overheads.}
Finally, we evaluate tracing and \texttt{Optimizer} overheads using the PyTorch pipelines of each domain (TF pipelines show similar results).
Table~\ref{tbl:tracing} reports the throughput overheads introduced by tagging each sample with metadata (to ensure correctness) and periodically tracing samples with statistics (100ms throughout our experiments).
While the overheads are slightly larger for high samples/s pipelines (i.e., NLP), these operations introduce minimal overall overheads.
}

\rev{
The \texttt{Optimizer} can also quickly explore optimizations.
As discussed in Section~\ref{sec:optimizer_static}, the \texttt{Optimizer} prunes the search space after each optimization pass, limiting the amount of plans it must evaluate.
For the \texttt{CV}, \texttt{SSD}, \texttt{NLP}, and \texttt{ASR} PyTorch pipelines, the \texttt{Optimizer} considered 251055, 5689, 101954, and 22741 plans, respectively.
The \texttt{Optimizer} was able to generate a solution in $<6$ seconds in each case, insignificant compared to long-running training jobs.
}

\section{Related Work}
\noindent\textbf{ML Input Data Frameworks and Optimizations.}
PyTorch DataLoaders~\cite{website:torch-utilsdata} and tf.data~\cite{vldb14:murray_tfdata} are native frameworks for PyTorch and TensorFlow, respectively. %
DataLoaders offer certain options for performance tuning such as multiprocessing and pinned memory, but requires manual configuration.
TorchData~\cite{website:torchdata} is a beta PyTorch data loading library and provides similar primitives to \texttt{Pipes}.
Its DataLoader2~\cite{website:dataloader2} is an incomplete prototype with an API for distributed processing, but active development has unfortunately stopped~\cite{website:torchdata-future}.
Meanwhile, tf.data can statically fuse and vectorize TF-native pipelines and optimize the CPU and RAM allocation to each operator.
Plumber~\cite{mlsys22:kuchnik_plumber} extends tf.data to use a linear program for resource allocation.
These frameworks only support local processing, limiting their ability to mitigate data stalls.

DPP~\cite{isca22:zhao_dsi} and GoldMiner~\cite{sigmod23:zhao_goldminer} are proprietary distributed services deployed at Meta and Alibaba, respectively.
Ray Data~\cite{website:ray-data} is an input data library built on top of Ray~\cite{osdi18:moritz_ray}.
Ray Data distributes processing using Ray's Task and Actor primitives, and optimizes for task overheads by fusing operators via fixed rules.
tf.data service~\cite{socc23:audibert_tfdataservice} offloads processing to distributed workers, but cannot support non-TensorFlow UDFs~\cite{website:tfdata-service-limitations}.
FastFlow~\cite{vldb16:um_fastflow} extends tf.data service to split processing between local and remote workers at a coarse granularity. %
\rev{Pecan~\cite{atc24:graur_pecan} is a concurrent work, built on top of tf.data service (and is thus applicable to only TensorFlow pipelines), that also studies transformation ordering, but does not support its concurrent application alongside other optimization passes.}
These systems, like \systemname, use an auto-scaling policy to tune worker parallelism.

Various systems extend input data frameworks to address orthogonal concerns.
Cachew~\cite{atc22:graur_cachew} extends tf.data service to create a service for multi-tenant environments, scaling processing and sharing cached samples between training jobs.
Cachew can identify ideal cache locations, but requires users to explicitly insert autocache operators, hindering its compatibility with other optimizations (e.g., reordering).
Cachew relies on tf.data's underlying optimizer, which we evaluated in Section~\ref{sec:evaluation}.
PRESTO~\cite{sigmod22:isenko_presto} is a profiler that determines the ideal location to cache, but requires users to manually implement suggestions.
CoorDL~\cite{vldb14:mohan_coordl}, OneAccess~\cite{hotcloud19:kakaraparthy_oneaccess}, Quiver~\cite{fast20:kumar_quiver}, Tectonic-Shift~\cite{atc23:zhao_tectonicshift}, and SiloD~\cite{eurosys23:zhao_silod} provide distributed caches for data shared across training jobs.

NVIDIA DALI~\cite{website:dali} and FusionFlow~\cite{vldb17:kim_fusionflow} use GPUs for input data processing.
Revamper~\cite{atc21:lee_refurbish}, SHADE~\cite{fast23:khan_shade}, and iCACHE~\cite{hpca23:chen_icache} introduce optimizations that modify input data semantics (e.g., up-sampling) to improve model convergence and input data efficiency.
RecD~\cite{mlsys23:zhao_recd} deduplicates recommendation datasets for input data processing efficiency.
\systemname's extensibility allows it to easily adopt these engines and techniques alongside its current optimizations.

As discussed in Section~\ref{sec:optimizations_requirements}, \systemname addresses the key requirements that are not met by current input data systems -- systematic, context-aware, and general optimizations.
\systemname provides an easy-to-use programming interface, supporting general ML frameworks and pipelines, that permits users to express a simple set of constraints that unlock a rich set of reordering and caching optimizations.
Meanwhile, its \texttt{Optimizer} systematically applies a complex set of optimizations to improve performance, and its \texttt{Scaler} leverages these benefits to efficiently meet training demand.

\noindent\textbf{Traditional Processing Frameworks.}
Naiad~\cite{sosp13:murray_naiad}, Spark~\cite{nsdi12:zaharia_spark}, DryadLINQ~\cite{osdi08:yu_dryadlinq}, and many other processing frameworks~\cite{cacm08:dean_mapreduce, vldb8:akidau_dataflow, sigmod06:meijer_linq, eurosys07:isard_dryad, jvldb14:alexandrov_stratosphere, icde11:borkar_hyracks} allow users to chain together higher-order transformations and model computation as data flows.
\rev{
Other data processing frameworks, such as Apache Beam~\cite{vldb8:akidau_dataflow, website:beam}, Apache Wayang~\cite{sigmod23:beedkar_wayang}, and RHEEM~\cite{vldb:kruse_rheem}, leverage an optimizer and decouple dataflow graphs from underlying execution engines, similar to \systemname's \texttt{Execution} interface.
\systemname is specialized for ML input data pipelines.
}

\systemname applies similar rule- and cost-based optimization passes to traditional query optimizers~\cite{icde93:graefe_volcano, dataengbull95:graefe_cascades, sigmod15:armburst_sparksql}, while also considering unique properties to ML input data pipelines such as relaxed order dependencies and randomness.
Lara~\cite{vldb12:kunft_lara} is a domain-specific language that optimizes matrix-heavy preprocessing for traditional ``shallow'' ML models (e.g., regressions).
Hueske \textit{et al.}~\cite{vldb5:hueske_blackbox} explore preserving UDF semantics under reordering by statically analyzing PACT~\cite{socc10:battre_pacts} programs.
\systemname extends the impact of these traditional data processing techniques to ML input data systems.
\section{Conclusion}
We presented \systemname, a unified framework to define, optimize, and execute ML input data pipelines.
\systemname's \texttt{Feature} API allows users to define input data pipelines using modular \texttt{Pipes} and to express lightweight hints that allow \systemname to reason about operator semantics.
\systemname automatically applies systematic, context-aware, and general optimizations to improve performance, and it orchestrates pipeline execution to efficiently meet training throughput demands. %
\systemname outperforms tf.data, tf.data service, FastFlow, Plumber, Ray Data, and PyTorch DataLoader by \rev{up to $1.87\times$ to $10.65\times$} across diverse pipelines.
\systemname provides extensible and general programming, optimizer, and execution interfaces that allow it to enable and evolve alongside future ML input data systems research.

\begin{acks}
We gratefully acknowledge Johann Hauswald, Andrew Woen, and our anonymous reviewers whose feedback has greatly helped improve this paper.
This research was partly supported by the Stanford Platform Lab and its affiliates, and by ACE, one of the seven centers in JUMP 2.0, a Semiconductor Research Corporation (SRC) program sponsored by DARPA.
Mark Zhao was supported by a Stanford Graduate Fellowship and a Meta PhD Fellowship.
\end{acks}

\balance
\bibliographystyle{ACM-Reference-Format}
\bibliography{paper}

\end{document}